\documentclass[final,5p,times,twocolumn]{elsarticle}
\usepackage{lineno,hyperref}
\usepackage{multirow}
\usepackage{multicol}
\usepackage{booktabs}
\usepackage{enumitem}
\usepackage{tikz}
 
\modulolinenumbers[5]

\journal{Journal of \LaTeX\ Templates}

\begin{document}

\begin{frontmatter}

%\title{Developing Future Human-Centered Smart Cities: \\Critical Analysis of Smart City Security, Interpretability, Data Management, and Ethical Challenges}

\title{Adversarial Attacks and Defenses for Social Network Text Processing Applications: Techniques, Challenges and Future Research Directions}

\author[1]{Izzat Alsmadi}
\ead{ialsmadi@tamusa.edu}
\author[2]{Kashif Ahmad}
\author[3]{Mahmoud Nazzal}
\author[4]{Firoj Alam}
\author[2]{Ala Al-Fuqaha}
\author[3]{Abdallah Khreishah}
\author[5]{Abdulelah Algosaibi}
\address[1]{Department of Computing and Cyber Security, Texas A\&M University, San Antonio, Texas, USA}
\address[2]{Information and Computing Technologies (ICT) Division, College of Science and Engineering (CSE), Hamad
Bin Khalifa University, Doha, Qatar.}
\address[3]{Department of Electrical and Computer Engineering, Newark College of Engineering, New Jersey Institute of Technology, Newark, NJ 07102, USA}
\address[4]{Qatar Computing Research Institute, HBKU, Doha, Qatar}
\address[5]{Department of Computer Science , College of Computer Science and Information Technology, King Faisal University, Saudi Arabia}
\cortext[cor1]{Corresponding Author}

%\address[5]{Affiliation }


\begin{abstract}
The growing use of social media has led to the development of several Machine Learning (ML) and Natural Language Processing (NLP) tools to process the unprecedented amount of social media content to make actionable decisions. %Examples include NLP tools for opinion mining, emotion recognition, and disinformation detection. The advancement of deep learning has also facilitated the development of more of such tools to process noisy social media content. Such tools are the building blocks for monitoring social media, reducing the spread of disinformation, and enabling business capacities.
However, these ML and NLP algorithms have been widely shown to be vulnerable to adversarial attacks. %Along this line, the vulnerability of NLP models in a variety of NLP applications has been established. With the important role of text in social media platforms, this vulnerability enables adversaries to target a variety of social media applications. 
%This allows adversaries to take part in social media Communions, affect public opinion, and manipulate the commercial activities on social media platforms. 
These vulnerabilities allow adversaries to launch a diversified set of adversarial attacks on these algorithms in different applications of social media text processing. In this paper, we provide a comprehensive review of the main approaches for adversarial attacks and defenses in the context of social media applications with a particular focus on key challenges and future research directions. In detail, we cover literature on six key applications, namely (i) rumors detection, (ii) satires detection, (iii) clickbaits \& spams identification, (iv) hate speech detection, (v) misinformation detection, and (vi) sentiment analysis. We then highlight the concurrent and anticipated future research questions and provide recommendations and directions for future work. 
\end{abstract}


\end{frontmatter}

%\linenumbers

\section{Introduction}
\label{sec:introduction}

Social media has become a major communication channel in everyday life. According to the Datareportal report published on 27th January 2021 \footnote{\url{https://datareportal.com/reports/digital-2021-global-overview-report}} more than half of the world population (4.48 billion) use social media. Among the social media platforms, the most popular ones include Facebook, Youtube, WhatsApp, Instagram, WeChat, and Twitter. 
% In the modern world, online social networks, such as Facebook, Instagram, and Twitter, is turned into a useful means of communication and 
Such platforms are widely used for information dissemination and consumption \cite{ahmad2019social}. Nowadays, social media platforms form an important gateway for connecting people, spreading thoughts, and linking business entities to customers through question advertisement, review collection, and feedback \cite{bhanot2012use}. The key advantage of such platforms is the wide scope of the audience that allows individuals to directly connect and share content. For example, expressing an opinion, sentiment, and emotion towards products, entities, individuals, and/or society \cite{liu2012sentiment,rout2018model,hassan2019sentiment,khan2019implicit}. 

The reach to a large audience and freedom of generating and sharing their content also poses several challenges. For instance, these platforms have also been used for political or financial gains (e.g., Brexit) \cite{GORODNICHENKO2021103772}, which has been typically done by targeting focused communities. Malicious actors have also been using them to spread rumors and mis/disinformation \cite{wu2019misinformation,alam2020fighting}. These platforms has also been used a ground for hate speech \cite{chetty2018hate,matamoros2021racism}, racism \cite{hasanuzzaman-etal-2017-demographic,onabola2021hbert}, xenophobia \cite{del2019sinai}, and prejudice \cite{vidgen2020detecting}. 
 
Two evolving major factors influence the growing attacks on machine learning algorithms and applications: (1) The rapidly growing dependence on automated decisions, and (2) The level and nature of influence such attacks can cause or trigger.
In Natural Language Processing (NLP) in general and Online Social Networks (OSNs) in particular, perhaps the Russian Troll Tweets (RTT) dataset and the influence in the USA 2016 election is one major example. RTT showed a recent trend in cyber attacks where humans, rather than their computing devices are targets of such attacks. Social bots and trolls are examples of text generators in Generative Adversarial Networks (GAN) models. The discriminators are first the social network websites and their efforts to eliminate social bots/trolls and counter the spread of mis/disinformation. The other category of discriminators is other users of social networks who are the targets of those bots/trolls and their media campaigns.
Autocratic countries see social network websites such as Twitter and Facebook as threats or weapons, \cite{farnsworth2011china}. This explains why many of those countries have state-sponsored agencies that are active in creating social media content for propaganda and anti-propaganda. They create social bots or trolls to target rival media. They also have social bots or trolls to target their own citizens. The widespread of social bots can be observed in recent social and political movements such as the Arab Spring, 2013-2014 protests in Ukraine, etc.

To address such issues a major research field has emerged, namely computational social science \cite{lazer2009life,conte2012manifesto}, a major part of it is to analyze social media content. The focus is automatically analysis of social media content using Machine Learning (ML) and Natural Language Processing (NLP) techniques. A major interest came from NLP research communities, which led to organizing workshops \cite{socialnlp-2021-international,socialnlp-2020-international,ws-2019-abusive,alw-2020-online}, evaluation campaigns \cite{shaar-etal-2021-findings,clef-checkthat-lncs:2020,clef-checkthat-en:2020,clef-checkthat:2021:LNCS,SemEval2021-6-Dimitrov,zampieri-etal-2019-semeval,zampieri-etal-2020-semeval}.
% Thanks to the recent development in Artificial Intelligence (AI), Machine Learning (ML), Natural Language Processing (NLP) technologies, social media content can be automatically processed to filter out unwanted content. 
% For instance, several tools have been developed to automatically check the authenticity of the information, and detect and remove rumors and misinformation from social networks \cite{song2019ced}. 
These efforts have made significant progress in the field in terms of automatically identifying such content, debunking them, and facilitating organizations, social media platforms, and society as a whole. 

While such progresses are in place, the discovery of \textit{adversarial examples} \cite{szegedy2013intriguing,goodfellow2014explaining} raised a major concern in different research communities and has been studied for many different problems including image classification \cite{moosavi2016deepfool,nguyen2015deep}, security \cite{papernot2016distillation}, malware detection \cite{grosse2017adversarial,suciu2019exploring}, robot-vision system \cite{melis2017deep}, medical diagnosis system \cite{finlayson2019adversarial,finlayson2018adversarial}, mis/disinformation \cite{wu2019misinformation,alam2020fighting}, fake-news \cite{le2020malcom}, and rumor \cite{10.1145/3308558.3313741}. 
%However, the proposed methods and developed tools are also subject to several adversarial attacks \cite{ahmad2020developing}. 
An adversarial attack refers to the process of crafting an adversarial example, which is intended to fool a model through adversarial perturbation. Adversarial attacks can be launched with different intentions including getting influence on a classifier's decision as well as violating security. One recent example of such attacks is the exploitation of the vulnerabilities of Microsoft’s Tay chatbot, which was shut down due to the racist tweets \cite{CHATBOT}. 

In the context of social media content analysis, the attackers (i.e., the individuals who want to misuse social media and spread inappropriate information) can perturb their content to ditch the AI-based content filters incorporated with social media platforms. For instance, an adversarial example may fool a hate speech detection system, which can result in misclassification of toxic and abusive content as acceptable and legitimate content. 
%\cite{}. 

% \todo[inline]{Can we point out which method has been proposed for which type of problem and type of content (e.g., text, image)}\fa{
To address the problem of adversarial attacks on ML models, there have been efforts, reported in the literature, to overcome such challenges. Such efforts include adversarial training with noise \cite{jenni2019stabilizing}, gradient masking \cite{vivek2019regularizer}, defense distillation \cite{wang2020defending}, ensemble adversarial learning \cite{tramer2017ensemble}, and feature squeezing \cite{xu2017feature}.
% }

% Several defense methods have been proposed in the literature for different types of content literature also reports several interesting defense methods to guard against adversarial attacks. To this aim, different strategies have been employed. Some key strategies include adversarial training with noise \cite{jenni2019stabilizing}, gradient masking \cite{vivek2019regularizer}, defence distillation \cite{wang2020defending}, ensemble adversarial learning \cite{tramer2017ensemble}, and feature squeezing \cite{xu2017feature}.

% The adversarial ML has been well explored on images \cite{moosavi2016deepfool,nguyen2015deep}. There are also significant efforts on adversarial ML on text \cite{} and speech \cite{}. 
%%The text-based adversarial examples are also called paraphrasing attacks and aim to modify the sequences of characters and words in text streams to fool the machine learning models. 
% In this paper, we mainly target adversarial attacks and the corresponding defense methods for text analysis in social networks.\footnote{Throughout the paper, we use social networks and social media interchangeably.} We particularly focus on key applications of social networks, such as {\em(i)} rumors, {\em(ii)} satires, {\em(iii)} click baits, {\em(iv)} spam, {\em(v)} hate speech, and {\em(vi)} misinformation.

\begin{table*}[t]
\centering
\scalebox{1.0}{
\begin{tabular}{p{2.0cm}p{2.5cm}p{9cm}}
\toprule
\multicolumn{1}{l}{\textbf{Ref.-Year}} & \multicolumn{1}{l}{\textbf{Domain}} &\multicolumn{1}{l}{\textbf{Main Focus}} \\ \midrule
\cite{belinkov2019analysis} 2019 & NLP & Adversarial attack methods in NLP. However, it does not address defense approaches. \\ \midrule

\cite{wang2019towards} 2019 & DNNs & Adversarial attacks and defense techniques for DNN-based solutions for English and Chinese text. The adversarial attacks are categorized on the perturbation units. The key text classifications covered in the survey include NLP inference, reading comprehension systems, questions and answer systems, and machine translation. \\ \midrule
\cite{zhang2020adversarial} 2020 & DNNs & Adversarial textual examples generation methods for DNNs. The key text analysis tasks discussed include machine translation, comprehension, text summarization, and text entailment. Also discusses multi-modal attacks on key applications of text to image, and image to text, such as scene-text recognition, image captioning, and visual question answering. \\ \midrule

\cite{xu2020adversarial} 2020 & Generic & Adversarial attacks and defense techniques for images, graph, and text analysis. \\ \midrule
\textbf{This Survey} 2021 & NLP in Social Media Applications & Adversarial attacks and defense systems for text-based social media applications, such as rumors, satires, clickbaits \& spam, hate speech, misinformation detection and sentiment. \\ \bottomrule
\end{tabular}
}
\caption{A comparative summary and of existing surveys.}
\label{tab:existing_surveys}
\end{table*}

\subsection{Scope of the Survey}
The adversarial ML has been well explored for images \cite{moosavi2016deepfool,nguyen2015deep}. There are also significant efforts on adversarial ML on text \cite{zhang2020adversarial} and speech \cite{grondahl2018all}, \cite{moh2020no}, \cite{khieu2019tsar}. In this paper, we mainly target adversarial attacks and the corresponding defense methods for text based social media applications.\footnote{Throughout the paper, we use social networks and social media interchangeably.} We particularly focus on key applications of social networks, such as {\em(i)} rumors, {\em(ii)} satires, {\em(iii)} click baits and spam, {\em(vi)} hate speech, {\em(v)} misinformation and {\em(v)} sentiment.
% The paper mainly focuses on adversarial learning, attacks, and defense methods proposed for text-based social media applications. The paper covers four important applications of social networks' textual analysis. These applications include (i) satires parodies and rumors detection, (ii) click baits and spam detection, (iii) hate speech detection, and (iv) identification of misinformation in social networks. 
Since text is a primary communication means in social media platforms \cite{farzindar2015natural}, the paper also provides detailed taxonomies of text adversarial attacks and corresponding defense techniques, which makes it self-contained in terms of relevant concepts. The paper also advises on key challenges, limitations, and future research directions for text-based social media applications. 

\subsection{Related Surveys}
The literature on adversarial text analysis is quite rich \cite{wang2019towards,zhang2020adversarial,xu2020adversarial}. 
%%It also reports some interesting surveys exploring different aspects of adversarial text analysis. 
Want et al. \cite{wang2019towards} provide an overview of the literature on adversarial attacks and corresponding defense strategies for DNN-based English and Chinese text analysis systems. Zhang et al. \cite{zhang2020adversarial}, provide a more detailed survey of adversarial attacks on deep learning-based models for NLP with a particular focus on adversarial textual examples generation methods. Xu et al. \cite{xu2020adversarial} on the other hand review the literature on adversarial attacks and defenses on images, graphs, and text analysis models. In contrast to existing surveys, this paper focuses on adversarial attacks and defense solutions for text-based social network applications. 
%%by covering literature on key social networks applications, such as satires, parodies \& rumors, clickbaits \& spam, hate speech, and misinformation detection. 
In Table \ref{tab:existing_surveys}, we provide an overview of existing surveys on adversarial learning, attack and defense methods.

\subsection{Contributions}
The key contributions of the survey can be summarized as follows.

\begin{itemize}
 \item We first provide a detailed taxonomy of adversarial attacks and the corresponding defense techniques for text analysis frameworks.
 \item We then provide a comprehensive survey of adversarial learning, attacks, and defense systems for major social media applications including rumors, satires, clickbaits \& spam, hate speech, misinformation detection, and sentiment.
 \item We also highlight key research challenges, open issues, and future research directions.
\end{itemize}

The rest of the paper is organized as follows. Section \ref{sec:taxonomy_attacks} provides a detailed taxonomy of adversarial attacks on the text analysis model. Attack techniques at the character, word, sentence, and inter levels are revised in Section \ref{sec:aa_nlp}. Section \ref{sec:taxonomy_defense} covers the corresponding defense techniques. Section \ref{sec:appications} provides an overview of adversarial attacks and corresponding defense solutions for different social media applications. Section \ref{sec:challenges} describes the key challenges and future research directions in adversarial text analysis. Finally, Section \ref{conclusion} provides some concluding remarks. 

\section{Adversarial Attacks}
\label{sec:taxonomy_attacks}

\begin{figure}[t]
\centering
\resizebox{0.99\columnwidth}{!}{
\includegraphics[width=0.45\textwidth]{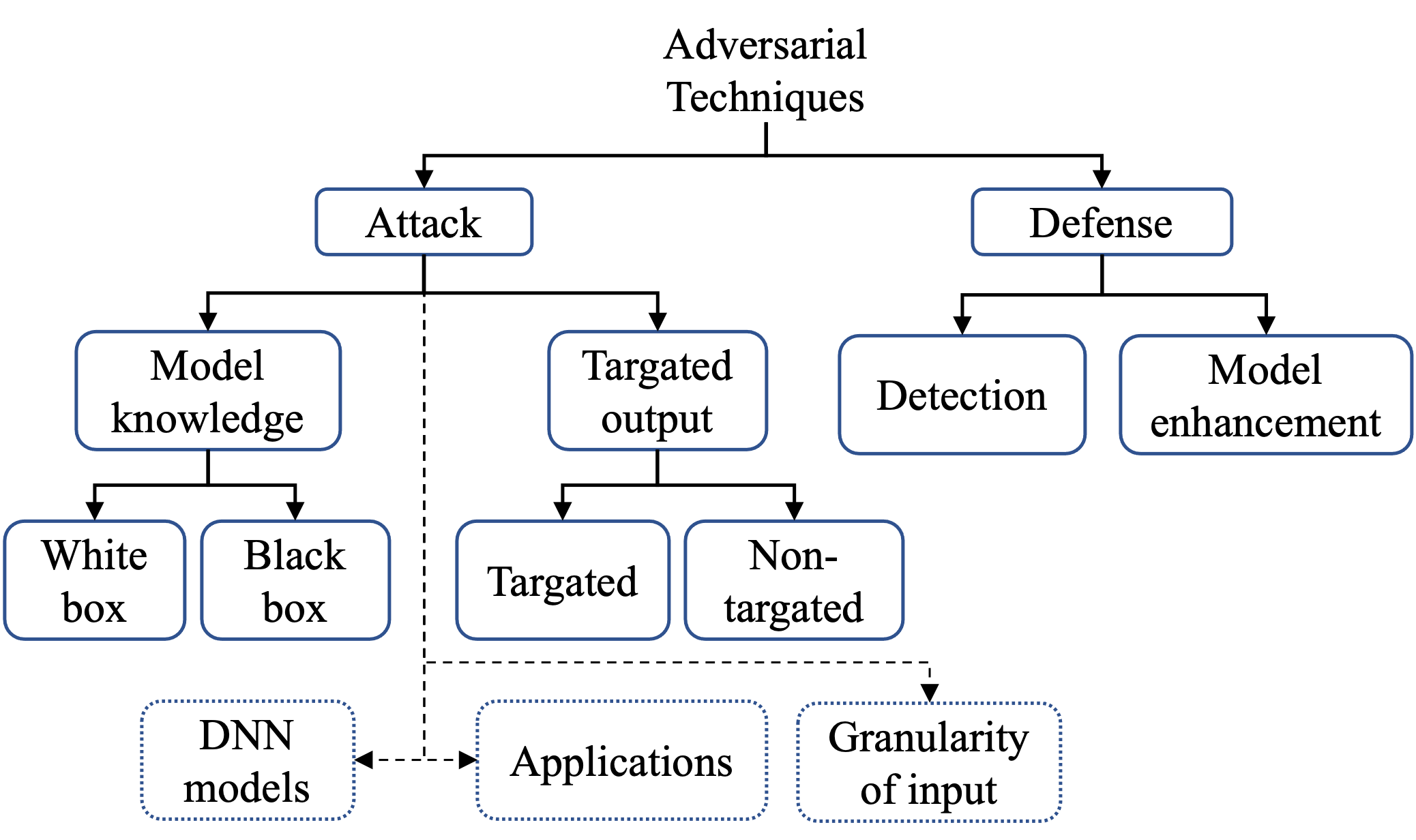}}
 \caption{Categorization of adversarial attack and defense techniques. Dotted boxes and arrows present that loosely defined categories.}
\label{fig:adver_tech_categorization}
\end{figure}

Current literature and several surveys categorized adversarial techniques into attack and defense \cite{zhang2020adversarial,wang2019towards}. These techniques can be further classified based on how attack and defense are applied to ML models. Such a classification have evolved based on the type of data (e.g. image, text), the knowledge level of the attacker (e.g. white vs. black box), the goal of the attack (targeted vs. non-targeted), and the granularity of the input content (e.g., for NLP, character, word, sentence). Other classifications include the modality of the input content, the type of application or task, and the architecture of the ML model\cite{zhang2020adversarial}. In Figure \ref{fig:adver_tech_categorization}, we list several categorizations of the main adversarial attack and defense techniques. In the figure, we use dotted lines and boxes to represent the categories that are loosely defined in the overall adversarial attack and defense techniques. For example, both model and target-based attacks apply to any model and application. Here is a brief description of each of these categorization criteria.

\textbf{Attack:}
The purpose of an adversarial attack is to manipulate the inputs and or the ML model so that it is fooled to produce faulty outcomes. This process takes on a variety of possible objectives and approaches.

\textbf{Model Knowledge:}
This category corresponds to classifying the attack techniques based on the attacker's knowledge about the target model. Broadly speaking, this can be sub-categorized into:
\begin{itemize}
 \item \textbf{White Box Attack:} In a white-box attack, the adversary knows and exploits the input and the model information to optimize its attack. The model information includes input-output data, architecture, parameters, loss, and activation functions. In this attack type, adversarial data can be generated in a way that maximizes its impact on the classifier while being an imperceptible change. Typically, an adversary adjusts adversarial modifications to be in the direction of the model’s gradient concerning the current input. This maximizes the increases in the loss function and thus optimizes the attack. White-box attacks have received a lot of attention in the literature due to their adaptive nature. 
 \item \textbf{Black Box Attack:}
 The black box attack techniques do not require or have access to any knowledge of the model except input and output. This type of attack uses heuristic approaches or repeated queries to build the attack. 
\end{itemize}

\textbf{Targeted Output:}
This feature categorizes attacks into targeted and untargeted. In a targeted attack, the adversary maps the original model's output to a required faulty output for a given input. However, in an untargeted attack, the adversary cares only about causing the model to produce incorrect outputs, regardless of what they might be.

\textbf{Defense:}
The goal of defense techniques is to design a robust model to fight against those attacks or to deal with such types of adversarial examples. As presented in Fig. \ref{fig:adver_tech_categorization}, the major defense techniques include {\em(i)} adversarial example detection and {\em(ii)} model enhancement. The goal of the former approach is to detect an adversarial example that is distinguishable from legitimate input. Whereas the goal of the latter approach is to train models with additional parameters, which are commonly referred to as adversarial training. 

Our focus in this study is social medial applications with a particular emphasis on NLP. Hence, in the following sections, we provide in-depth studies of adversarial attacks and defense techniques reported in social media applications.

\section{Adversarial Attacks in NLP}
\label{sec:aa_nlp}
Today's social media platforms use a versatile bundle of networking tools, such as messaging, chatting, voice, and video calling. Amongst these, text is a primary tool for communication in social media \cite{farzindar2015natural}. Therefore, we will especially focus on the research body on adversarial attacks and defenses on NLP techniques from a social media perspective.

Adversarial attack in an NLP context faces more challenges compared to the case of computer vision models. Primarily, the discrete nature of text inputs imposes certain limitations on how one can efficiently modify them. In this regard, character, word, and sentence-level modifications are primarily text replacements at the reception levels. To this end, many challenges face the replacement process. Examples include how one characterizes token \footnote{A token means a character or a word.} or sentence similarity, establishing the embedding space, i.e., the space of acceptable replacements, and more importantly, how to search for the best replacement candidate within such a space. For NLP tasks, adversarial attacks are mainly categorized based on the granularity of the input. The main categorization of such granularity of the input includes the following.

\begin{enumerate}
 \item Character-level attacks: achieved by adding, deleting, or swapping a character in a word.
 \item Word-level attacks: achieved by replacing an original word with an adversary equivalent. There is a variety of methods to characterize this word \textit{equivalency}. Examples include synonyms, antonyms, and semantic equivalence.
 \item Sentence-level attacks: are carried out by adding, deleting, or paraphrasing certain sentences. 
 \item Inter-level attacks are combinations of character, word, or sentence level attacks. 
\end{enumerate}

\begin{figure}[t]
\centering
\resizebox{0.97\columnwidth}{!}{
\includegraphics{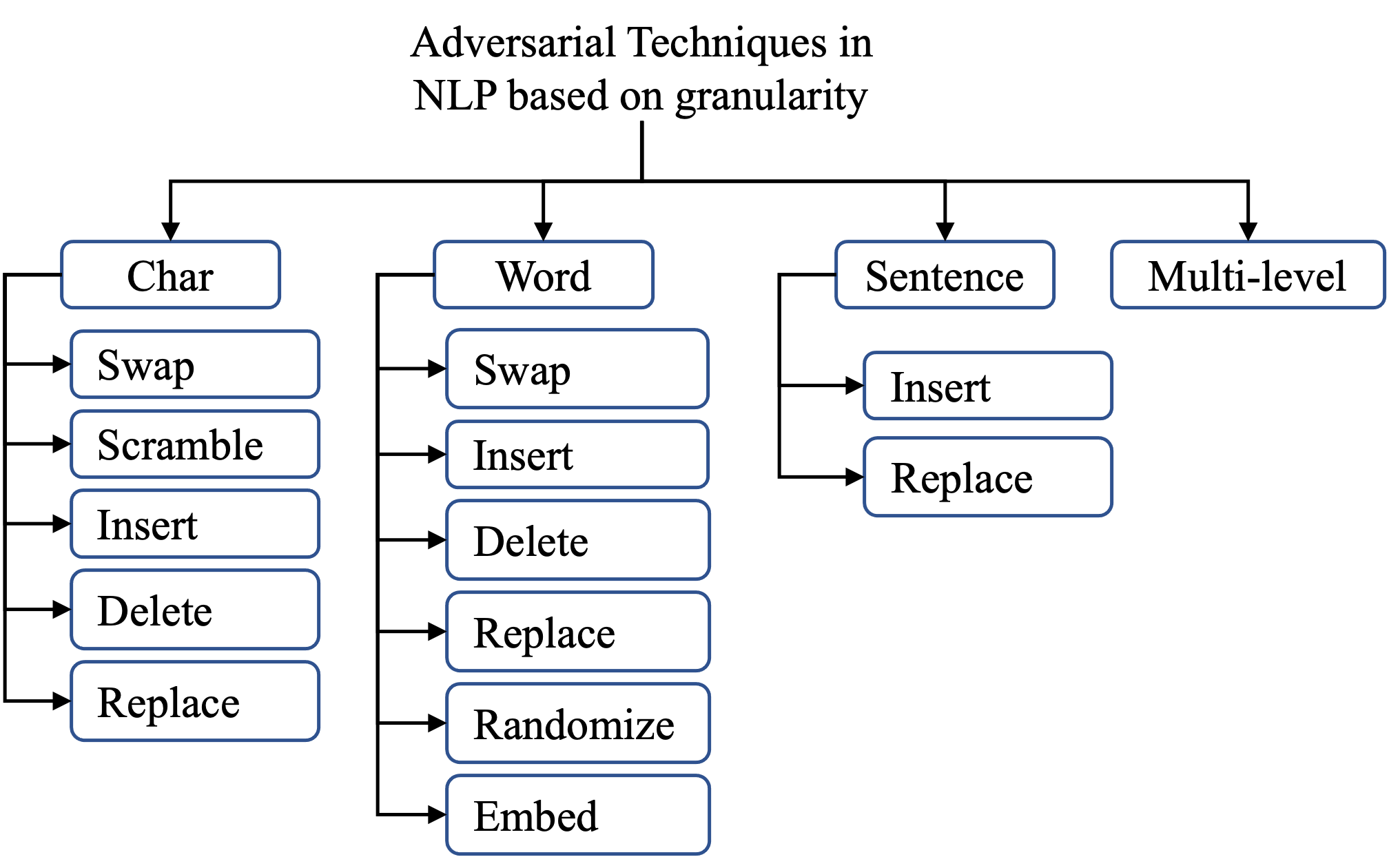}}
 \caption{Categorization of adversarial attack for NLP tasks.}
\label{fig:nlp_adver_tech_categorization}
\end{figure}

Such categorizations are inspired based on how these linguistic components are used in NLP applications to train the machine learning model. For example, character and word level n-grams and their different representations (e.g., tf-idf, word2vec, contextual representation using transformers) have been used to train the machine. In Figure \ref{fig:nlp_adver_tech_categorization}, we present the most common categories of granularity-based adversarial attacks used in NLP tasks. 

\begin{table}[htbp]
\centering
\scalebox{1.0}{
\setlength{\tabcolsep}{2.0pt}
\begin{tabular}{@{}llllll@{}}
\toprule
\multicolumn{1}{c}{\textbf{Ref.}} & \multicolumn{1}{c}{\textbf{Swap}} & \multicolumn{1}{c}{\textbf{Scramble}} & \multicolumn{1}{c}{\textbf{Delete}} & \multicolumn{1}{c}{\textbf{Insert}} & \multicolumn{1}{c}{\textbf{Replace}} \\ \midrule
\cite{ebrahimi2017hotflip} & N & N & Y & Y & Y \\
\cite{gao2018black} & Y & N & Y & Y & Y \\
\cite{li2018textbugger} & Y & N & Y & Y & Y \\
\cite{belinkov2017synthetic} & Y & Y & N & N & Y \\ \bottomrule
\end{tabular}
}
\caption{Character modifications of the main \textbf{character-level adversarial attacks}. Y: Yes, N: No.}
\label{tab:adaptive_adversarial_attacks}
\end{table}

\begin{table}[htbp]
\centering
% \resizebox{\textwidth}{!}{%
\scalebox{1.0}{
\setlength{\tabcolsep}{2.0pt}
\begin{tabular}{@{}lllll@{}}
\toprule
\textbf{Ref. } & \textbf{\begin{tabular}[c]{@{}l@{}}Model \\ Acc.\end{tabular}} & \textbf{\begin{tabular}[c]{@{}l@{}}Targeted \\ attack\end{tabular}} & \textbf{\begin{tabular}[c]{@{}l@{}}Targeted \\ Model\end{tabular}} & \textbf{\begin{tabular}[c]{@{}l@{}}Targeted \\ Application\end{tabular}} \\ \midrule
\cite{ebrahimi2017hotflip} & W & N & CharCNN & TC \\
\cite{gao2018black} & B & N & CharCNN & SA, TC \\
\cite{li2018textbugger} & W, B & N & CharCNN & TC \\
\cite{belinkov2017synthetic} & B & N & CharCNN & MT \\
\cite{hosseini2017deceiving} & B & Y & Perspective & Toxic word detection \\
\cite{brown2019acoustic} & B & N & Perspective & Toxic word detection \\
\cite{pruthi2019combating} & B & N & DNN & SC, paraphrase detection \\
\cite{eger2019text} & B & N & CharCNN & Multiple \\
\cite{le2020malcom} & W, B & N & DNN & Fake news detection \\ \bottomrule
\end{tabular}%
}
\caption{Relevant work for the main \textbf{character-level adversarial attacks} in terms of model knowledge, targeted attack, targeted model, and applications. B: black-box, W: white-box, Y: yes, N: no, CharCNN: character-level CNN, DNN: deep neural network, TC: text classification, SA: sentiment analysis, MT: machine translation, SC: sentiment classification. Model, and, Acc.: Model Accessibility.}
\label{tab:character_level_adversarial_attacks}
\end{table}

\subsection{Character-level Attacks}

Character level information has been widely used in NLP for text classification tasks. \cite{zhang2015character} presents a pioneering work on applying convolutional neural networks (CNN) for text classification at the character level. This work has paved the way for character-level operations. In Table \ref{tab:adaptive_adversarial_attacks}, we summarize several character-level perturbations. These primarily include insertion, deletion, substitution/replacing, swapping, and/or scrambling a character. In Table \ref{tab:character_level_adversarial_attacks}, we list sample works on character-level attacks with their model accessibility, attack type, targeted model, application, or task. 

As a pioneering work on the character-level attack, \cite{ebrahimi2017hotflip} investigates white-box attack with character-level adversarial examples
to maximize the model's loss at limited numbers of modifications. This is referred to as the HotFlip algorithm. 
It is based on performing an atomic flip operation where the gradient of the model is used to select between several perturbations with respect to a one-hot vector representation. Such perturbations can be obtained by flipping, inserting, or deleting a character. In this regard, the authors argue that a simple beam search approach is superior to a greedy search between the flip operations. 

Along the line of black-box attacks is the work of \cite{gao2018black}, referred to as DeepWordBug based on characters and words. Despite its black-box nature, this method is adaptive to the input. The key idea is to adaptively choose the most (\textit{critical)} tokens. DeepWordBug operates in two main stages. \textit{First}, it identifies critical tokens, which is based on the ranking of the perturbed tokens measured in terms of the classifier’s output. \textit{Second}, it changes the identified tokens using simple
transformations. Such transformations include four primary types; {\em(i)} swapping two consecutive characters, {\em(ii)} substituting a character with another one by randomly selecting from the same word, {\em(iii)} randomly deleting a character, and {\em(iv)} inserting a randomly selected character. This study opened a new research direction in terms of \textit{what to modify and why?}. Subsequently, this question was addressed by the TextBugger method \cite{li2018textbugger} that considered the more general framework of deep learning-based text understanding. Thus, TextBugger mainly differs in the way how the tokens are ranked. Similar to DeepWordBug, it operates in two successive stages. \textit{First}, it identifies the most dominant tokens for modification. \textit{Second}, it modifies the identified tokens. TextBugger selects between five modifications; {\em(i)} inserting a space between characters, {\em(ii)} randomly deleting a character, {\em(iii)} swapping two characters selected at random, {\em(iv)} replacing a character with another visually similar character, and {\em(v)} replacing a word with a semantically similar one. 

\subsection{Word-level Attacks}
\begin{figure*}[htb]
\centering
\includegraphics[width=0.75\textwidth]{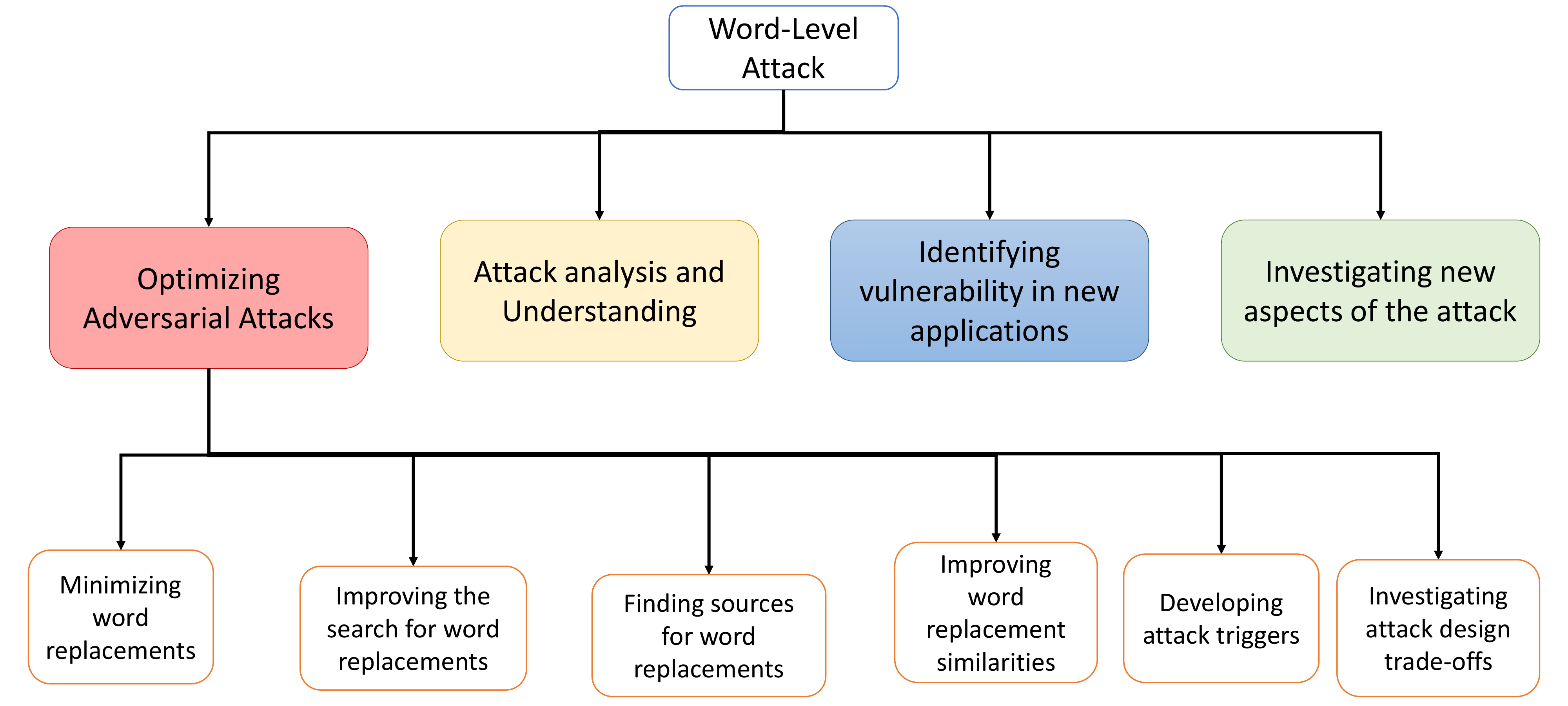}
\caption{A summary of the main research contributions in the area of \textbf{word-level attack}.}
\label{word_level_contribution_tax}
\end{figure*}

Compared to character-level attacks, a word-level attack is naturally more imperceptible for humans and more difficult for machine learning algorithms to defend. In Table \ref{tab:word_level_adver_attacks_modification}, we summarize several word-level perturbations proposed in several research works. In Table \ref{tab:word_level_adversarial_attacks}, we report several word-level attack contributions and categorize them based on model accessibility, attack type, targeted model, application, or task. In a broader sense, summarize the main research contributions made at word-level adversarial attack in Fig.~\ref{word_level_contribution_tax}. These can be divided into four main categories; {\em(i)} optimizing adversarial attacks, 
{\em(ii)} identifying vulnerability in new applications, {\em(iii)} attack analysis and understanding, and {\em(iv)} investigating new attack aspects. Below we discuss each of them in detail. 

\begin{table}[]
\centering
% \resizebox{\textwidth}{!}{%
\scalebox{1.0}{
\setlength{\tabcolsep}{2.0pt}
\begin{tabular}{@{}llllllll@{}}
\toprule
\textbf{Ref.} & \textbf{Insert} & \textbf{Delete} & \textbf{Swap} & \textbf{Flip} & \textbf{Sub.} & \textbf{Rand.} & \textbf{Embed} \\ \midrule
\cite{ebrahimi2017hotflip} & N & N & N & N & Y & N & N \\
\cite{ebrahimi2017hotflip} & Y & Y & Y & Y & N & N & N \\
\cite{kuleshov2018adversarial} & N & N & N & N & Y & N & N \\
\cite{jin2020bert} & N & N & N & N & Y & N & N \\
\cite{gao2018black} & Y & Y & Y & N & Y & N & N \\
\cite{zang2020word} & N & N & N & N & Y & N & N \\
\cite{sato2018interpretable} & N & N & N & N & Y & N & N \\
\cite{zhou2019learning} & Y & Y & N & N & N & Y & Y \\
\cite{ren2019generating} & N & N & N & N & Y & N & N \\ \bottomrule
\end{tabular}%
}
\caption{Word modifications of the main \textbf{word-level adversarial attacks}. Sub.: Substitution, Rand.: random attack randomly samples a word to replace the target word. \textbf{Embed}: it replaces the word with a word among the top-10 nearest words in the embedding space. 
%(1 and 0 denotes Yes and No, respectively.)
}
\label{tab:word_level_adver_attacks_modification}
\end{table}

\begin{table}[!thb]
\centering
\scalebox{.85}{
\setlength{\tabcolsep}{2.0pt}
% \begin{tabular}{@{}p{0.05\textwidth}lllll@{}}
\begin{tabular}{@{}lllll@{}}
\toprule
\multicolumn{1}{l}{\textbf{Ref.}} & \multicolumn{1}{l}{\textbf{\begin{tabular}[c]{@{}c@{}}Model\\ Access\end{tabular}}} & \multicolumn{1}{l}{\textbf{\begin{tabular}[c]{@{}c@{}}Targeted \\ attack\end{tabular}}} & \multicolumn{1}{l}{\textbf{\begin{tabular}[c]{@{}c@{}}Targeted\\ Model\end{tabular}}} & \multicolumn{1}{l}{\textbf{\begin{tabular}[c]{@{}c@{}}Targeted\\ Application\end{tabular}}} \\ \midrule
\cite{papernot2016distillation} & W & N & DNN & IC \\
\cite{ebrahimi2017hotflip} & W & Y & CharCNN-LSTM & TC \\
\cite{ebrahimi2017hotflip} & W & N & CharCNN-LSTM & MT \\
\cite{gao2018black} & B & N & WordLSTM, CharCNN & SA, TC \\
\cite{kuleshov2018adversarial} & W, B & N & LSTM, CNN & TC \\
\cite{yang2020greedy} & B & N & CharCNN, WordCNN & TC \\
\cite{jin2020bert} & B & N & CNNS, RNNs & TC, TE \\
\cite{zang2020word} & B & N & DNN & TC \\
\cite{wallace2019universal} & W & N & DNN & TC, MRC and CTG \\
\cite{garg2020bae} & B & N & WordCNN, WordLSTM & TC \\
\cite{ribeiro2018semantically} & B & N & DNN & MRC, SA, VQA \\
\cite{alzantot2019genattack} & B & N & LSTM & SA, TE \\
\cite{li2016understanding} & B & N & DNN & Sequence tagging,WC \\
\cite{lei2018discrete} & B & N & WordLSTM, CharCNN & SA \\
\cite{sato2018interpretable} & W & N & LSTM & SC, CC,GEC \\
\cite{zhou2019learning} & B & N & BERT & TC \\
\cite{ren2019generating} & B & N & CNN, LSTM & TC \\ 
\cite{wang2019bilateral} & W & N & DNN & IC \\
\cite{wang2020towards} & N.A. & N & DNN & Object detection \\
\cite{gong2018adversarial} & W & N & CNN & TC, SA \\
\cite{alzantot2018generating} & W & Y & CNN & TC, SA \\
\cite{liang2017deep} & W, B & Y & CNN & TC \\

\bottomrule
\end{tabular}
}
\caption{Relevant work for the main \textbf{word-level adversarial attacks}. B: black-box, W: white-box, Y: yes N: no, MRC: machine reading comprehension, word-level CNN: WordCNN, word-level SLTM: WordLSTM, TC: text classification, MT: machine translation, SA: sentiment analysis, SA: sentiment classification, TE: text entailment, IC: image classification, GEC: grammatical error classification, WC: word classification, CTG: conditional text generation. Tar.: Targeted.}
\label{tab:word_level_adversarial_attacks}
\end{table}

\subsubsection{Optimizing adversarial attacks}
A major part of research on a word-level attack is focused on optimizing adversarial attacks. In this study, we categorize the endeavors made in optimizing word-level attacks into the following sub-categories.

\paragraph{Minimizing word replacements}
The first research body under the umbrella of optimizing world-level attacks focuses on minimizing word replacements. Generally speaking, the most common perturbation type in word-level attack is based on word replacement. Typically, a target word is replaced by an equivalent one chosen from a space of possible replacements. Such a space was identified by Ebrahimi et al. \cite{ebrahimi2018adversarial} and referred to as the embedding space. In \cite{ebrahimi2018adversarial}, the authors consider neural machine translation (NMT) and propose elementary modifications to achieve word-level adversarial attacks. In this regard, the authors extend the HotFLip algorithm proposed in \cite{ebrahimi2017hotflip} to the word level. Essentially, Hotflip is assumed to have white-box access. Thus it uses the model's gradient to optimize word modification selection such that the model's loss is maximized greedily. Similar to other works, word modifications include adding, removing, or replacing a word in the translation input. 

Another work on a word-level adversarial attack is proposed in \cite{alzantot2018generating}. In this work, the authors highlight the differences between adversarial attacks in vision and NLP domains. In particular, they point out that one can not arbitrarily modify text inputs as done in a computer-vision context. Thus, the most important contribution in this work is necessitating the importance of exploiting semantic and syntactic similarity between an original word and its adversarial replacements. Such similarities were overlooked by the literature before this work. The authors exploit these similarities in word replacement search using a population-based optimization algorithm to generate the embedding space of adversarial samples exhibiting semantic and synthetic similarities. Besides, they use this algorithm to select the best replacement of a given original word with respect to semantic and syntactic similarity and maximizing the model's loss. It is noted that they selected the word to be modified at random. Technically, this process is conducted in the following order.
\begin{enumerate}
\item Compute the nearest neighbors for a given word within its embedding space based on Euclidean distance.
\item Rank the similarity of those neighbors to the original word in terms of the difference between their label productions and that of the original word. Then keep a few neighbors having the maximum similarity
\item Amongst the neighbors of maximum similarities, select the one that maximizes the loss function of the model.
\end{enumerate}
%\noindent The authors reported achieving a 100\% success by altering an average of 8.7\% of the words in the given text in their experiments.

\begin{figure}[]
\centering
\resizebox{0.99\columnwidth}{!}{
\includegraphics[width=0.35\textwidth]{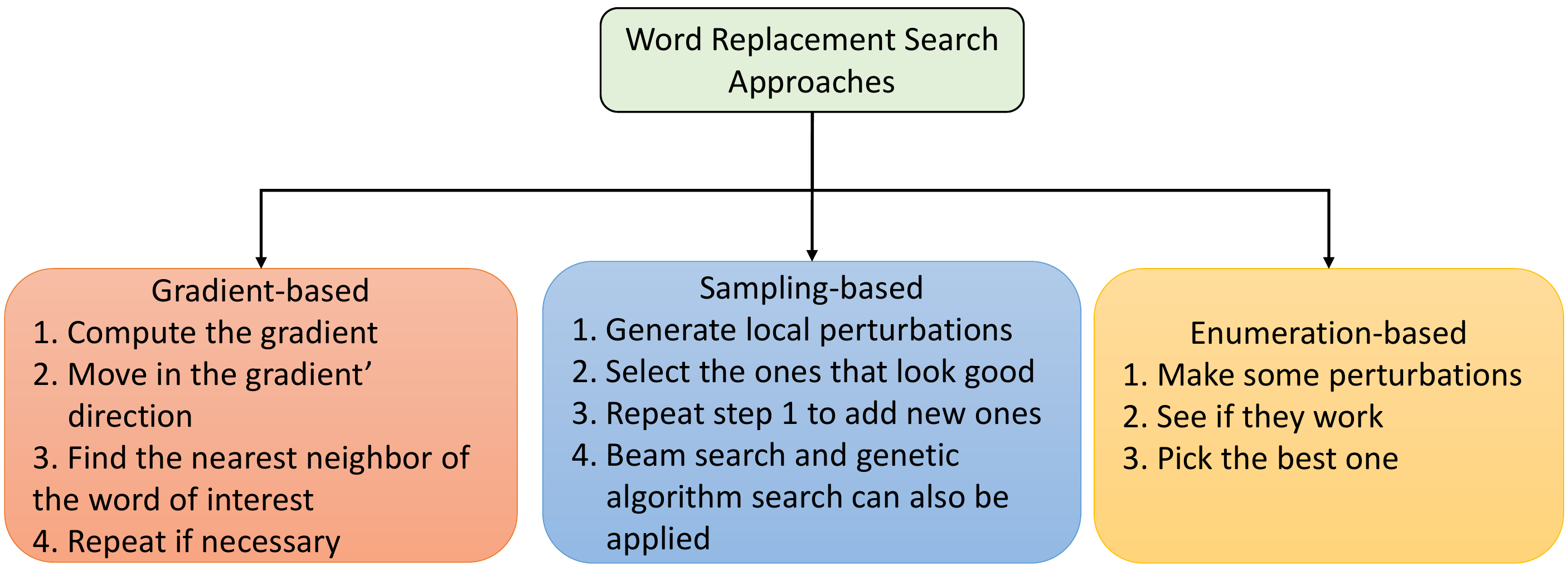}}
\caption{Word replacement search techniques.}
\label{search_approaches}
\end{figure}

\paragraph{Improving the search for word replacements}
Another sub-category of works on optimizing adversarial word-level attacks is concerned with improving the search for word replacements in an embedding space.
In the current literature, there are three main categories of text replacement techniques: {\em(i)} gradient-based \cite{ebrahimi2018adversarial,wu2019misinformation,gong2018adversarial}, {\em(ii)} sampling-based \cite{papernot2016distillation,alzantot2019genattack}, and {\em(iii)} enumeration based \cite{belinkov2017synthetic,iyyer2018adversarial,ribeiro2018semantically}, as shown in Fig. \ref{search_approaches}. These techniques vary in the extent of the assumed model accessibility, and thus the information used to dictate the search process. 

Herein, we provide two examples of improving the search process in the embedding space. First, the above-mentioned DeepWordBug algorithm \cite{gao2018black} also applies to the word level. In this setting, the same selection process is adopted. In other words, the algorithm works in two stages; first, it selects a set of critical words and then chooses the best amongst them. Another example is the TextBugger algorithm of \cite{li2018textbugger} which also applies to the word level. 

\paragraph{Obtaining embedding spaces}
 
\par As discussed earlier, selecting the best replacement is of crucial importance for the success of the attack. Still, it is also important to first provide a suitable embedding space. So, the next research body under the umbrella of optimizing word-level attacks concerns obtaining the embedding space. Along this line, \cite{alzantot2019genattack} propose a black-box, gradient-free genetic optimization algorithm for generating adversarial replacements based on natural selection of the genetic algorithm. Specifically, it is an iterative process for providing a generation of selections per iteration. It is noted that a fitness function is used to quantify the quality of a candidate. Besides, candidates of each generation are obtained by a cross-over of their preceding ancestors. 

\par Several approaches aim at leveraging methods for defining image-based embedding space to the text application area while keeping in mind the differences between text and image data. Along this line, \cite{gong2018adversarial} proposes a framework for utilizing ideas from image attacks to be used as text attacks. First, the authors highlight the main differences between image and text domains in terms of generating adversarial attacks. In this regard, the authors report the bottleneck in this borrowing which can be cast as two main limitations. First, is the discrete nature of text rendering small perturbations inapplicable in a text context. Second, is the difficulty in evaluating how good an adversarial attack is. As a remedy to the first limitation, the authors propose searching within the set of possible word embedding models and then choosing the best adversarial alternative to the original word in terms of its nearest neighbor similar to the original. To alleviate the second limitation, the authors use a Word Mover’s Distance (WMD) measure as an adversarial replacement quality metric. 

\paragraph{Improving word replacement similarities}

\par It is well-known that an adversarial replacement should be, at least visually, \textit{similar} to the original word. Moreover, it is necessary to quantify this similarity. Therefore, another research body under the umbrella of optimizing word-level attacks focuses on improving the similarities used in identifying word equivalency. Along this line, \cite{ribeiro2018semantically} introduces two approaches for quantifying word similarity. First, is the concept of semantically equivalent adversaries (SEAs). In this context, SEAs are used to generate semantically equivalent adversarial replacements based on paraphrase generation techniques. These techniques include back-translation, i.e. translating a given sentence to a certain language, and then back-translation of the original language. Another example is the use of paraphrasing to obtain such replacements. In this setting, it is noted that the semantics of the text are preserved since the meaning is not changed. The second approach proposed by the authors is devising equivalent adversarial rules (SEARs) that govern the process of obtaining semiannually equivalent replacements.

\par In line with identifying and exploiting word replacement similarity, Ren et al. \cite{ren2019generating} consider imposing further restrictions on adversarial replacements. Namely, lexical correctness, grammatical correctness, and semantic similarity. Specifically, the authors propose an algorithm for adversarial word selection ordering where the saliency of the word and its classification probability are taken into account for determining its order in the embedding space. This algorithm is referred to as probability-weighted word saliency (PWWS) for text adversarial attacks. Ren et al. experimentally validate a high success rate of their algorithm at minimal word replacements. Moreover, the authors demonstrate that their modifications are salient at the human level through human evaluation of the attacks. Similar to other methods, this algorithm also possesses a high degree of transferability of its examples across different models and datasets.

\par To this end, word-level adversarial replacement techniques are accused to tend to overlook the linguistic context while replacing targeted words. This renders them vulnerable to human discovery. Accordingly, \cite{garg2020bae} calls for a wiser selection of the words. In particular, the authors incorporate context consistency as a constraint for replacement in a black-box adversarial attack algorithm. Algorithmically, this is achieved by sorting replacement suitability values based on a BERT-masked language model. 

\paragraph{Developing attack triggers}
\par The studies surveyed so far consider primarily input-dependent adversarial attacks. Therefore, the next research body under optimizing word-level attack considers developing (input-agnostic) attack triggers. Along this line, \cite{wallace2019universal} investigates input-agnostic adversarial examples, referred to as triggers, at the token level in NLP models. The authors propose a gradient-based search for the best token to change and use minimal perturbation length. In \cite{wallace2019universal}, the authors validate a high success rate of their algorithm as tested over various tasks and networks. Besides, they demonstrate that it possesses a degree of transferability regardless of its white-box nature. In this regard, such attacks transfer across both examples and models for all tasks. As a future outlook, the authors necessitate investigating the development of grammatical triggers that can work anywhere in the input. Besides, future research may also consider dataset- or even task-agnostic triggered. Moreover, the authors report that adversarial attacks raise questions on whose responsibility it is that models are vulnerable to such attacks. 

\paragraph{Investigating attack design trade-offs}
\par The next research body under the umbrella of word-level attack concerns identifying and balancing design trade-offs. An inherent trade-off in the design of adversarial attacks is the saliency-strength trade-off. In this context, a strong perturbation is better to change the model output at the cost of reduced saliency, and vice-versa. This forms the fundamental design trade-off. However, identifying and balancing other design trade-offs was not considered until the work of \cite{yang2020greedy}. In this work, the authors devised and investigated a probabilistic framework for generating adversarial attacks on models with discrete inputs. \cite{yang2020greedy} propose two methods for world-level adversarial attacks and then investigate the underlying design trade-offs encountered with these methods. 
. 
\subsubsection{Attack analysis and understanding}
\par The second aspect of research contributions in word-level attacks concerns developing the understanding of adversarial attacks. In this regard, \cite{li2016understanding} argues that the interest in the applicability of ML models came at the cost of not giving enough care to the interpretation and the understanding of how ML models yield their outputs based on inputs. To bridge this gap, the authors investigate the sensitivity of the model’s operation to input deletion, in particular. Specifically, the authors analyze how the behavior of a given model will change with erasure operations of different types and natures on input data. The authors investigate the use of reinforcement learning for detecting output-changing phrases at the input side. The result of the work conducted by \cite{li2016understanding} is a set of important interpretations and explanations of certain ML operational phenomena. 

\subsubsection{Identifying vulnerability in new applications}

\par Another research body in the context of word-level attack focuses on identifying the vulnerability to adversarial attacks exhibited by new NLP tasks and application areas. In this regard, Ling et al. \cite{liang2017deep} consider the vulnerability of DNN-based text classifiers to adversarial input attacks. In particular, the authors propose three main methods for generating text adversarial examples; insertion, removal, and modification. Experiments validate that these methods, whether applied on the character level or the word level, can trick DNN models and cause them to generate wrong outputs. This is while the perturbations generated by such methods are still imperceptible at the human level. 

\par It is worth mentioning that Liang et al.'s work is based on identifying the best characters to change, i.e., the \textit{hot} characters. Then, it applies adversarial changes on other words containing more than 3 hot characters and refers to them as \textit{hot words}. Similarly, \textit{hot phrases} are identified as the ones having the maximum number of hot words. Therefore, this approach seems conceptually similar to what HotFlip does.

\par Another work by \cite{kuleshov2018adversarial} considers the vulnerability of natural language classification models to adversarial attacks. This work establishes the existence of adversarial attacks in an NLP context. Specifically, it points out that, similar to other ML application areas, natural language classification models are vulnerable to attacks composed of small perturbations at the input leading to changing the NLP model's output. 

\par As an attempt at investigating other text-related ML applications, \cite{wang2020towards} consider ML models used in the application of object detection. The authors show that such models are, similar to the case with other ML applications, inherently vulnerable to adversarial attacks on the image patch level. This is shown by proposing an efficient adversarial attack algorithm that targets a specific class of the object detection model. Thus, the authors craft their attack by tracking the model to be blind with respect to a certain object in an image patch called the invisibility patch. This algorithm has the ability to track several state-of-the-art object detection approaches. Besides, it transfers across different models and datasets. 

\subsubsection{Investigating new aspects of the attack}
\par Lastly, another research body of word-level attack contributions focuses on the attack itself. The common practice in adversarial machine learning has remained restricted to altering the input/model while leaving the input labels intact. In contrast to this widely-used assumption, \cite{wang2019bilateral} proposes a bilateral perturbation process, i.e., altering both input and the label. Thus, for altering a given input's label, the authors propose a closed-form mathematical expression. However, to alter an input, the authors use a one-step perturbation process while adopting a class label that maximizes confusion. The idea of perturbing both input and the label has been shown effective. 

%%%%%%%%%%%%%%%%%%%%%%%%%%%%%%%%%%%%%%%%%%%%%%%%%%
%%%%%%%%%%%%%%%%%%%% Sentence-level %%%%%%%%%%%%%%
%%%%%%%%%%%%%%%%%%%%%%%%%%%%%%%%%%%%%%%%%%%%%%%%%%

\subsection{Sentence-level attacks}
In sentence-level attacks, an adversary considers the inputs as sentences, and thus, applies adversarial modifications on them by inserting, replacing, or deleting sentences. Compared to character-and word-level attacks, sentence-level attacks demand a longer time in adversary text generation. 
The key attributes of several sentence-level adversarial attacks are presented in Table \ref{tab:sentence_level_adversarial_attacks}. 
% Below is a detailed account of several contributions made in this research area. 

\begin{table}[htbp]
\centering
\scalebox{1.0}{
\setlength{\tabcolsep}{2.0pt}
\begin{tabular}{@{}lllll@{}}
\toprule
\textbf{Ref.} & \textbf{\begin{tabular}[c]{@{}l@{}}Model \\ Acc.\end{tabular}} & \textbf{\begin{tabular}[c]{@{}l@{}}Targeted \\ attack\end{tabular}} & \textbf{\begin{tabular}[c]{@{}l@{}}Targeted \\ Model\end{tabular}} & \textbf{\begin{tabular}[c]{@{}l@{}}Targeted \\ App\end{tabular}} \\ \midrule
\cite{lei2018discrete} & W & N & WCNN, LSTM & TC \\
\cite{jia2017adversarial} & W, B & N & DNN & QA \\
\cite{iyyer2018adversarial} & B & N & LSTM & SA, TE \\
\cite{cheng2019robust} & W & N & Transformer (\cite{vaswani2017attention}) & MT \\
\cite{michel2019evaluation} & W & N & LSTM, Transformer (\cite{vaswani2017attention}) & MT \\
\cite{jethanandani2020adversarial} & B & N & DNN & Lip reading \\
\cite{zheng2020understanding} & W, B & N & DNN & IC \\ \bottomrule
\end{tabular}
}
\caption{Specifications of the main \textbf{sentence-level adversarial attacks}. B: black-box, W: white-box, QA: question answering, TC: text classification, MT: machine translation, SA: sentiment analysis, and TE:textual entailment.}
\label{tab:sentence_level_adversarial_attacks}
\end{table}

% \cite{lei2018discrete}

There are several challenges to conducting adversarial attacks on models with textual input (e.g., sentence). One is finding suitable candidate replacements so that generated text preserves syntax and semantics. Another is developing an efficient approach to finding good transformations. The work of Lei et al. \cite{lei2018discrete} proposes methods that address both challenges {\em(i)} sentence and word paraphrasing that preserve syntax and semantics, {\em(ii)} gradient-guided greedy paraphrasing approach to find suitable transformations. 
Along this line, Iyyer et al. \cite{iyyer2018adversarial} propose syntactically controlled paraphrase networks to generate a paraphrase of the sentence, which is syntactically valid. This study shows that such generated adversarial examples i) can fool pre-trained models, and ii) when trained as augmented data can improve the robustness of the model. 
The work of Jia and Liang \cite{jia2017adversarial} also explores sentence-level adversarial examples, where the authors investigated whether a distorted sentence in a paragraph can lead to an incorrect answer.

\subsection{Multi-level text adversarial attacks}
By multi-level attack, we mean an adversarial attack that perturbs characters and words or words and sentences, or all of these. A multi-level attack is largely considered unexplored and open for future research. The few research studies mentioned in this section evaluate a preliminary approach to integrating attacks at different text levels. 
Another reason why multi-level attacks are unexplored heavily in literature is that they have nothing comparable to image-based adversarial attacks. As adversarial attacks in text are more recent than adversarial attacks in images, many of the methods in text attacks are inspired by early image attacks.

As multi-level attacks involve at least two of the three text-level attacks, they are typically more complex and computationally expensive. 
Similar to the earlier categories, multi-level attacks can be used in white-box attacks (e.g., \cite{ebrahimi2017hotflip}, \cite{jia2017adversarial}, \cite{wallace2019universal}), as well as black-box attacks (e.g., \cite{blohm2018comparing}). 
% They can be also used for different applications such as general classification, question answering, etc). The main approach described in HotFlip white-box attack is at the character level. Additionally, researchers showed that their approach can be extended to word-level attacks as well. 

\section{Defense Methods in NLP}
\label{sec:taxonomy_defense}
\par Similar to the case of adversarial attacks, adversarial defense research in NLP is still in its early stages as compared to its counterparts in computer vision. In the broad sense, the existing efforts in this domain can be divided into four categories: {\em(i)} adversarial training, {\em(ii)} spell and grammatical check, {\em(iii)} defensive distillation, and {\em(iv)} recovery of perturbations.
In the next subsections, we provide a detailed overview of the literature on each type of adversarial defense methods. 

\subsection{Adversarial training}
Adversarial training is the process of training a model on a mixture of clean data and adversarial examples. The inclusion of adversarial examples in such a training set has been shown to improve the model's robustness against those examples\cite{goodfellow2014explaining}. This defense scheme aims at suppressing the impact of an adversarial attack, assuming it has already happened or will inevitably happen. Intuitively, the re-trained model is expected to be more robust to the adversarial attacks whose instants are included in the training, as compared to the original model. In the adversarial security research area, adversarial training has been used by almost all adversarial attack works \cite{benz2021universal}. It is worth mentioning that, improving the model robustness by adversarial training has been widely used as an implicit indicator of the success of the underlying adversarial attack. 

Despite being the first line of defense, there are certain limitations and drawbacks of adversarial training, as summarized below.
\begin{enumerate}
 \item Some recent works are skeptical about the extent to which adversarial training can truly improve the model's robustness. However, to address such limitations more robust approaches have all been proposed \cite{sato2018interpretable,pruthi2019combating}. Some works report marginal/moderate robustness improvement for the adversarial training in certain NLP applications \cite{sato2018interpretable,pruthi2019combating}. 
 \item Adversarial training has recently been shown effective exclusively with the attack whose adversarial samples are used in its training \cite{jia2017adversarial}. This led to the inception of ``blind-spot" attacks \cite{zhang2019limitations} where the modified input resides in a so-called blind spot. This refers to the state of being far away from regular and potential inputs while still belonging to the data distribution. Thus, any adversarial training attempt is very unlikely to include such an example in its training set. 
 \item The current adversarial training procedures cannot scale to datasets with a large (intrinsic) dimension \cite{zhang2019limitations}. Besides, adversarial training is shown to work well only with repetitive data patterns in a training set.
\end{enumerate}

\par Recently, there has been an interest in extending the classical adversarial training concept. Along this line, \cite{sato2018interpretable} considers an inherent gap between adversarial training in image processing and NLP application domains. While perturbations used in adversarial training for image processing can take any arbitrary direction, the authors signify that this approach should not directly apply to the NLP case. This is because such an approach ignores the inter-portability of the added perturbations. Therefore, the authors call for restricting the perturbation space to the dimension of possible word embedding models. In other words, perturbations in an NLP context should only yield meaningful tokens from a linguistic perspective. Through a set of experiments, the authors demonstrate that their approach generates interpretable adversarial outcomes while maintaining the attack performance. 

\subsection{Spelling and Grammatical Check}

As character or word-level attacks are based on modifying characters and words, several studies investigate the utility of spelling and grammatical checks of the input text as means for defense \cite{gao2018black,li2018textbugger,pruthi2019combating,liu2020co,bao2021defending}. There have been several contributions in this direction. For instance, 
Gao et al. \cite{gao2018black} use an auto spell corrector to boost and improve the model's robustness against adversarial attacks. Similarly, Li et al. \cite{li2018textbugger} employ spell check as a tool for eliminating character-level modifications. In a more recent work, \cite{pruthi2019combating} propose appending text classification by word recognition. This aims at counteracting the effect of adversarial perturbations applied to characters. In this regard, various perturbations and changes such as character swapping, replacement, and keyboard typos can be counter-acted. 

\par Word-level input correction defense considers context-guided spell and grammar checks to guard against adversarial attacks. For instance, in \cite{liu2020co}, a two-stage spell correction scheme to identify and correct misspelled words is proposed to guard against adversarial attacks. As another example, Bao et al. \cite{bao2021defending} propose a multi-task learning framework to identify adversarial sentences where adversarial words are embedded. 

\subsection{Defensive distillation}

The distillation concept originated as a means for reducing the size of a given DNN architecture. This is possible by the virtue of training a smaller network on the logits and the training data of a given (larger) network. Thus, the smaller network will inherit the same functionality as the former one. More recently, \cite{papernot2016distillation} has adopted this idea as a defensive means for combating adversarial attacks. Specifically, the authors anticipate that the trained network will only inherit the benign functionality of the original one while being \textit{distilled} from any adversarial functionality. It is noted that the above-mentioned approach has been shown successful in combating adversarial attacks in the image domain \cite{zhang2021adversarial}. 

\subsection{Recovery of Perturbations}
The literature also witnesses some efforts for the identification and recovery of perturbations. For instance, a defense framework proposed by Zhou et al. \cite{zhou2019learning} aims to determine whether a particular token is a perturbation or not. The process is composed of two phases. In the first phase, a set of possible perturbations that could have been applied to the text of interest is identified. The second phase aims at calculating the value of a specific attack estimator for each of the attack possibilities. After that, the token that is identified as a perturbation is reconstructed/restored from the embedding space based on a certain similarity measure. The authors used the k-nearest neighbors (kNN) to guide the search. The study has also concluded that the technique is generally successful in identifying attacks in a variety of NLP models and applications without the need for any retraining.

Another work in \cite{eger2019text} develops a rule-based recovery of adversarial perturbations. The defense technique is based on replacing each non-standard character in the input stream with its nearest standard neighbor. In other words, this approach is based on reversing a text attack. The proposed solution has been proven very effective in machine learning translation applications.

\section{Adversarial Attack and Defense on Social Media Applications}
\label{sec:appications}

In this section, we study five major social media applications that are vulnerable to adversarial attacks and summarize the literature that explores attack and defense techniques for these applications.

\subsection{Rumors}
\label{ssec:rumor}

%\section{Satires Parodies Rumors}
% \todo[inline]{Firoj}

% \paragraph{Rumor}: 
As social media has become a major communication channel where users are not only consuming news but also propagating them by sharing and producing them. Such a consumption and dissemination approach led to widespread rumors which causes serious consequences to the individuals, organization, or the society as whole \cite{10.1145/3161603}.
%Rumor diffusion and convergence during the 3.11 earthquake: a Twitter case study
To address such an issue social media platforms, government entities, journalists, fact-checkers, research communities, and other stakeholders are fighting to reduce the negative impact of rumors. While fact-checking organizations such as FactCheck.org and Snopes.com are manually checking and making them publicly available to facilitate other stakeholders and reduce the spread. As such an effort does not scale well, hence, the research community has been trying to develop an automated system to detect and alert users about such rumors (see recent surveys \cite{cao2018automatic,10.1145/3161603}). 

In literature \textit{rumor} has several definitions~\cite{cao2018automatic,cao2018automatic}. {\em(i)} Rumor is a story or statement whose truth value is yet to be verified at the time of posting \cite{allport1947psychology}. {\em(ii)} The truth value of a story or statement is verified authoritative sources and confirmed that it is false or fabricated, which is also referred to as \textit{false rumor} \cite{wu2015false}. {\em(iii)} The third definition of a rumor is based on the users' subjective judgment of the veracity of the story or statement \cite{castillo2011information}. The former one is the most widely used and consistent with the definition of different dictionaries (e.g., Oxford English Dictionary). 
%%A rumor can be understood as an item of information that has not yet been verified, and hence its truth value remains unresolved while it is circulating. A rumor is defined as unverified when there is no evidence supporting it or there is no official confirmation from authoritative sources (e.g., those with a reputation for being trustworthy) or sources that may have credibility in a particular context (e.g., eyewitnesses).

Rumor can be of different types: {\em(i)} long-standing rumors that have been circulating for a long time, {\em(ii)} emerging rumors during any event (e.g., rumor about COVID-19). Depending on the type of rumor classification systems have been designed accordingly. 

\begin{figure}[t]
\centering
\resizebox{0.97\columnwidth}{!}{
\includegraphics[width=0.45\textwidth]{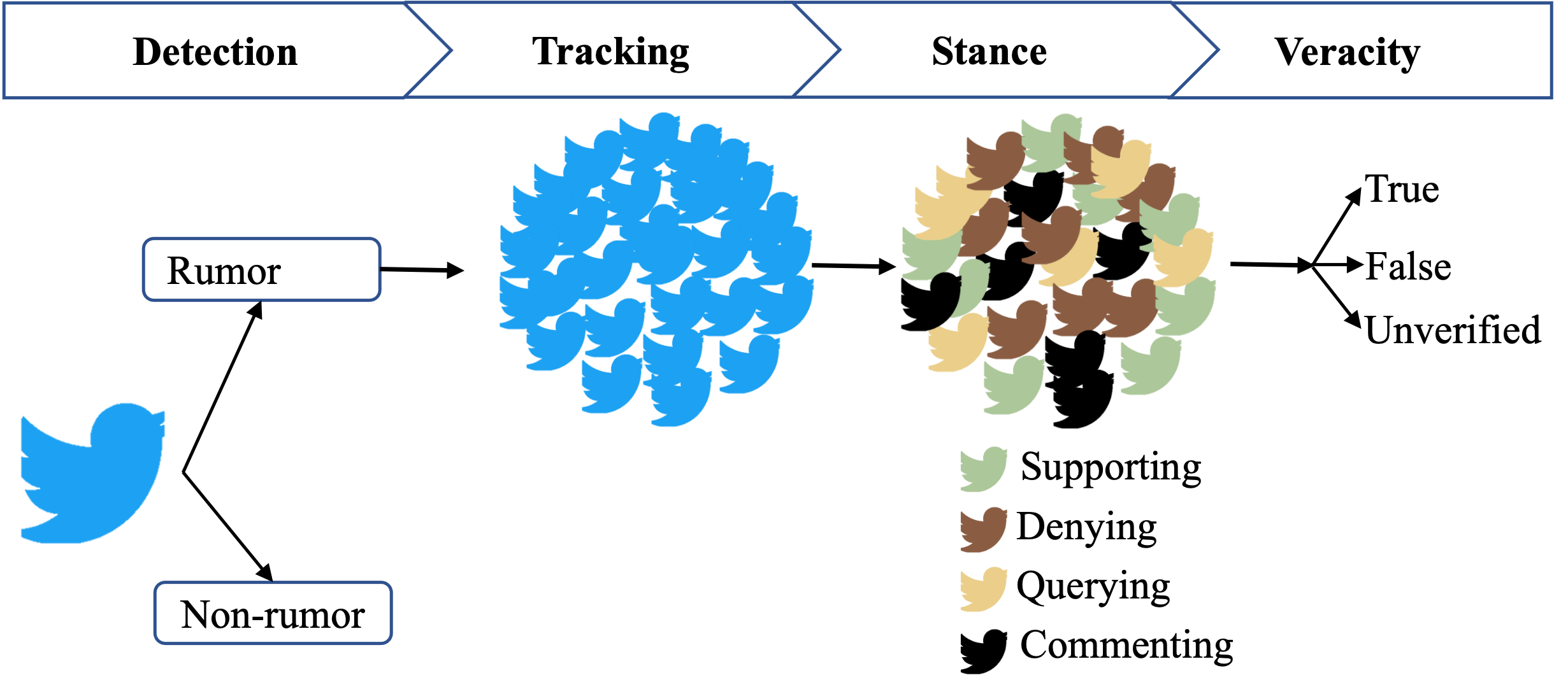}}
 \caption{A typical system architecture of rumor classification system. Adopted from \cite{10.1145/3161603}.}
\label{fig:rumor_detection_system}
\end{figure}

A typical rumor classification system consists of several components, as can be seen in Figure \ref{fig:rumor_detection_system} (adopted from \cite{10.1145/3161603}). News or social media posts are monitored to detect whether they are rumors or not. Once a rumor is identified then posts related to the rumor are detected to flag them. The stance classification component then determines how each post reflects a particular stance on its like veracity. Then the last component determines the truth value of the rumor.

To design the machine learning models for such components different types of methods have been used. The work by Cao et al. \cite{cao2018automatic} categorizes three types of methods. \textbf{Handcrafted feature-based approach:} features are extracted from different modalities (textual, visual, and social features) depending on their availability then classical machine learning methods (e.g., Decision Tree, Bayesian Networks, SVM, Random Forest, and Logistic regression) are used to train the model. \textbf{Propagation-based approach:} uses network-based information such as users, messages, and events to train the model. \textbf{Deep learning approach:} trains the model using a combination of different sets of features.

\textbf{Adversarial work:}
While malicious actors are trying to spread rumors to achieve political and financial goals, the research community and social media platforms are trying to develop automated systems/models to debunk them. These models are susceptible to malicious attacks too by malicious actors. Hence, it is important to develop models that are robust enough to deal with such adversarial attacks. Work in this direction for rumor detection are relatively very few. Ma et al. \cite{10.1145/3308558.3313741} is the first study that attempted to create robust models using GAN, which has been evaluated using two real-world datasets. 
Xiaoyu et al. \cite{ijcai2020-197} propose graph adversarial learning to graph structure to reduce the intentional perturbations. Song et al. \cite{song2021adversary} propose a method, which includes weighted-edge transformer-graph network, and position-aware adversarial response generator modules. In Table \ref{tab:soa_rumor_detection}, we report notable work on rumor detection, which propose different attack and defense methods.

\begin{table}[]
\centering
% \resizebox{\textwidth}{!}{%
\scalebox{0.78}{
\setlength{\tabcolsep}{2.0pt}
\begin{tabular}{@{}lllll@{}}
\toprule
\multicolumn{1}{c}{\textbf{Ref.}} & \multicolumn{1}{c}{\textbf{Dataset}} & \multicolumn{1}{c}{\textbf{Method}} & \multicolumn{1}{c}{\textbf{Attack}} & \multicolumn{1}{c}{\textbf{Defense}} \\ \midrule
\cite{10.1145/3308558.3313741} & \begin{tabular}[c]{@{}l@{}}Twitter \cite{10.5555/3061053.3061153},\\ Pheme \cite{10.1007/978-3-319-67217-5_8}\end{tabular} & GAN & \begin{tabular}[c]{@{}l@{}}Generator generates \\ proximately realistic\\ examples\end{tabular} & \begin{tabular}[c]{@{}l@{}}Discriminator tries \\ to distinguish \\ between real and fake\end{tabular} \\
\cite{ijcai2020-197} & \begin{tabular}[c]{@{}l@{}}Weibo \cite{song2019ced}, \\Twitter \cite{zubiaga2016learning}\end{tabular} & \begin{tabular}[c]{@{}l@{}}Graph \\adv. \\ learning\end{tabular} & \begin{tabular}[c]{@{}l@{}}Dynamically add \\ intentional perturbations \\ on the graph structure\end{tabular} & \begin{tabular}[c]{@{}l@{}}Learn more distinctive \\ structure features to resist \\ such perturbations\end{tabular} \\
\cite{song2021adversary} & \begin{tabular}[c]{@{}l@{}}Pheme \cite{10.1007/978-3-319-67217-5_8}, \\Twitter 15,\\ Twitter16 \cite{ma-etal-2017-detect}\end{tabular} & \begin{tabular}[c]{@{}l@{}}WETGN,\\ PRAG\end{tabular} & White box attack & \begin{tabular}[c]{@{}l@{}}Generates new adver-\\ sarial responses\end{tabular} \\ \bottomrule
\end{tabular}%
}
\caption{Recent work of adversarial techniques for \textbf{rumor detection}. WETGN: Weighted-Edge Transformer-Graph Network, PARG: Position-aware Adversarial Response Generator.}
\label{tab:soa_rumor_detection}
\end{table}

\subsection{Satires}
\label{ssec:satires}
% \todo[inline]{Firoj}
%% according to the Oxford dictionary, it is defined as ``The use of humor, irony, exaggeration, or ridicule to expose and criticize people's stupidity or vices, particularly in the context of contemporary politics and other topical issues.''
Satire is defined in the Cambridge Dictionary as ``a way of criticizing people or ideas in a humorous way''.\footnote{\url{https://dictionary.cambridge.org/}} It is a literary device that writers use to mock or ridicule a person, group, or ideology by judging them for various issues, particularly in the context of contemporary politics and other topical issues \cite{li2020multi}. Such devices include humor, irony, sarcasm, exaggerations, parody, or caricature \cite{mchardy2019adversarial}. These are typically applied to news and social media posts and the purpose is not to cause harm but to ridicule, or expose behavior that is shameful, corrupt, or otherwise ``bad'' \cite{tandoc2018defining,golbeck2018fake,10.1145/3395046}. Even the intention is not to mislead, however, the content can be mistaken by the reader as legitimate news, which can lead to the spread of misinformation \cite{li2020multi,garret2019too} and can be harmful \cite{yang-etal-2017-satirical}. Hence, it became essential to automatically identify them at a large scale and the typical classification task was to differentiate between real, fake, or satire news \cite{golbeck2018fake,li2020multi}.

For designing the classification model, different classical and deep learning-based algorithms have been used. These include Naive Bayes Multinomial, LSTM, CNN \cite{golbeck2018fake,burfoot-baldwin-2009-automatic}. To train the model typical features include 
entity mentions \cite{burfoot-baldwin-2009-automatic} coherence information \cite{goldwasser-zhang-2016-understanding}, distributions of parts-of-speech, sentiment, and exaggerations \cite{rubin-etal-2016-fake}. Other approaches include neural network and attention mechanism to incorporate paragraph-level linguistic features \cite{yang-etal-2017-satirical}, hierarchical deep neural network to incorporate sentence level and at the document level \cite{de-sarkar-etal-2018-attending} and multimodal learning \cite{li2020multi} using state-of-the-art vision-linguistic model ViLBERT.

\textbf{Adversarial work:}
The work on adversarial training for satire detection is relatively new. The study in \cite{mchardy2019adversarial} used adversarial training while training the model, however, the purpose was not to defend an adversarial attack, rather the idea was to control the effect of publication source information in satire detection. 

\subsection{Clickbaits and Spams}
\label{ssec:clickbaits_spams}

Social media platforms, such as Facebook and Twitter, have enabled people around the globe to connect and exchange ideas with each other. In doing so, they have also created opportunities for cyber-attackers and cyber-trolls to disseminate misinformation, offend and cyber-bully the users, and cause disputes through social media and public forums. \cite{jachim2020trollhunter} describe online trolls as malicious users pretending to be sincere members of a group discussion but subtly attempting to disrupt the discussion and cause conflict. \cite{tromble2019you} claim that the lack of attention to social science by the designers and developers of social media is responsible for spreading misinformation. Cyber-attackers leverage digitalization for online discrimination, privacy breaches, %false iatrogenesis, 
misinformation, and cyberattacks \cite{mckee2019second}.

\cite{cinelli2019mis} provides an example of misinformation in Italy. The Five Star Movement, founded by a national celebrity, linked with numerous social media pages and websites promoting the Catholic faith and spreading nationalist, anti-immigrant, and anti-Islam rhetorics. Similarly, \cite{tenove2019online} addresses the political threats posed by misinformation towards elections, democracy, and citizens via OSNs. These threats include releasing fake documents and ridiculing the candidates, spreading misinformation about the voting procedure, and harassing and intimidating minority groups. Also, the authors highlight the steps, such as extending existing criminal and human rights laws to online space, taken by the Canadian government to mitigate cyber-attacks.

ML models have proven effective in detecting malicious emails, fake news, and harmful data traffic in real-time (\cite{yaoartificial}). \cite{uppal2020fake} propose a method by drawing a connection between the surface features and the document-level characteristics of discourse for automated detection of fake news and misinformation. The proposed model achieves a veracity accuracy of 74\%. Likewise, \cite{gadekmanipulation} combines text-mining, behavior analysis, and connection graphs of a discourse, which involves users. It compares the results with a predefined knowledge base as an automated fact-checking and fake news detection model. 

Attackers can leverage adversarial machine learning (AML) to evade detection and amplify their attacks (e.g., an attack on spam email filter \cite{biggio2013evasion}). 
% \cite{wallace2019universal} Universal-Trigger significantly decreases the accuracy of ML detection models by generating a constant text and attacking it to a benign input. FA: this one does not fit here
\cite{jachim2020trollhunter} introduces TrollHunter, an ML model to detect the spread of misinformation and fake news on Twitter by leveraging linguistic analysis. TrollHunter achieves an accuracy of 98.5\% when detecting malicious tweets. The authors also present TrollHunter-Evader, which utilizes adversarial machine learning techniques, such as Test Time Evasion (TTE) and Ambient Tactical Deception (ATD), to evade detection, with a success rate of 40\%. \cite{le2020malcom} Malicious Comment Generation Framework (MALCOM) generates relevant and acceptable phrases. It replaces them with the ones in a malicious discourse to evade ML detection models with a success rate of 94\%. 

Cyber-attackers utilize various psychological and technology-based techniques to perform scams and hoaxes on their targets. Internet users face online threats in the forms of phishing, pharming, hacking, profiling, and physiological influence \cite{chiesa2008profiling}. In response, \cite{cabrera2019mail} implements an anti-spamming email filter based on unsupervised Artificial Neural Networks (ANNs), which achieves ROC AUC of 0.97, with an average sensitivity $>$ 0.95. However, the performance declines due to concept drift as emails with new topics and structures are introduced to the model. The accuracy of ML models decreases over time due to changes in news topics, the way they are reported, and evasion techniques used by adversaries, known as `concept drift' \cite{horne2019robust}. Thus, the authors recommend two approaches to mitigate the effects of concept drift: periodically training machine learning using the previous data samples and deploying Dynamic Weighted Majority into the system. 

In addition to adversaries causing misclassification by masking the input data with noise and poisoning training samples, implementation of ML models can be challenging due to their complexity and lack of transparency, making them incomprehensible to humans, making their decisions untrustworthy \cite{zhao2020event}. To mitigate the negative perception of AI, \cite{zhou2019fake} recommends educating developers on technology-based ethics and the potential effects on society, having a regulatory authority to monitor AI-based processes and making the AI computations and databases transparent to the consumers.

In Table \ref{tab:soa_clickbait_spam_detection}, we summarize the related work on clickbaits and spam detection addressing different attack and defense methods. 

%%%%%%%%%%%%%%%%%%%%%%%%%%%%%%%%%%%%%
 \begin{table}[]
\centering
% \resizebox{\textwidth}{!}{%
\scalebox{0.78}{
\setlength{\tabcolsep}{2.0pt}
\begin{tabular}{@{}lllll@{}}
\toprule
\multicolumn{1}{c}{\textbf{Ref.}} & \multicolumn{1}{c}{\textbf{Dataset}} & \multicolumn{1}{c}{\textbf{Method}} & \multicolumn{1}{c}{\textbf{Attack}} & \multicolumn{1}{c}{\textbf{Defense}} \\ \midrule

\cite{wallace2019universal}&GOSSIPCOP &RNNs &\begin{tabular}[c]{@{}l@{}} Universal \\targeted\\ attack \end{tabular}& \begin{tabular}[c]{@{}l@{}}Uses word\\ recognizer\\and topic \\relevancy \\method\end{tabular}\\ 
\cite{koenders2021vulnerable}& Untrue.News \cite{woloszyn2020untrue}&\begin{tabular}[c]{@{}l@{}}RoBERTa, \\ BERTweet,\\ FlairEmbeddings\end{tabular} & \begin{tabular}[c]{@{}l@{}} TextAttack \\ recipes \cite{morris2020textattack} \end{tabular}& N/A \\ 

\cite{kantartopoulos2020exploring}& &AdaBoost& Relabeling attack \cite{boatwright2018troll}& \begin{tabular}[c]{@{}l@{}} a k-NN-based \\ mechanism \\to sport \\outliers \end{tabular}\\ 
%& & & & \\ 
%& & & & \\ 
 \bottomrule
\end{tabular}%
}
\caption{Recent work of adversarial techniques for \textbf{clickbaits and spam detection}.}
\label{tab:soa_clickbait_spam_detection}
\end{table}
%%%%%%%%%%%%%%%%%%%%%%%%%%%%%%%%%%%%%

\subsection{Hate Speech Detection}
\label{ssec:hate_speech}

Hate speech, which refers to abusive or threatening speech or writing expressing a preconceived opinion against a particular group based on prohibitive attributes, such as color, race, religious beliefs, gender, or sexual orientation, is considered as one of the main causes of increasing global violence \cite{laub2019hate}. Social networks, such as Facebook, Instagram, and Twitter, and greater accessibility to Internet have further amplified it by allowing people to express their opinions to the global audience more freely and effectively \cite{chetty2018hate,matamoros2021racism}. 

Considering the increasing concerns over global violence, several efforts have been made to identify the potential sources and reduce the spread of hate speech over social networks. AI, ML, Data Science, and NLP communities are also playing their part by proposing interesting hate speech detection techniques \cite{schmidt2017survey}. For instance, \cite{mozafari2019bert} proposes a hate speech detection and classification framework for Twitter text streams based on BERT (Bidirectional Encoder Representations from Transformers).
. However, similar to other NLP applications, hate speech detection methods are also subject to adversarial attacks, for instance, as demonstrated in \cite{oak2019poster} where hate speech recognition models were fooled by modifying the text using state-of-the-art NLP adversarial attacks. The goal of such adversarial attacks is to disturb the classification capabilities of the classifiers resulting in misclassification of abusive and toxic content. The literature provides several examples of adversarial attacks on hate speech detection techniques. For instance, Grondahl et al. \cite{grondahl2018all} employed three different types of adversarial attacks to fool hate speech recognition models through {\em(i)} word changes, {\em(ii)} word-boundary changes, and {\em(iii)} appending unrelated innocuous words. Similarly, in \cite{moh2020no} hate speech detection models are fooled through {\em(i)} typos, {\em(ii)} removing white spaces, {\em(iii)} inserting benign words, and (iv) appending character boundaries. The authors also launched attacks on the models by combining all the individual types of attacks resulting in a more effective adversarial attack. 

To guard against the attacks on hate speech detection models, several interesting defense strategies have been introduced in the literature. The majority of the proposed methods rely on adversarial training to cope with the perturbations \cite{xu2020adversarial,tran2020habertor,xia2020demoting,grondahl2018all}. For instance, in \cite{tran2020habertor} adversarial training is used to guard against Fast Gradient Method (FGM) attack by extending FGM with a learnable and fine-grained noise magnitude where noise (perturbation) is added to misleading samples. Besides adversarial training, some solutions also rely on pre-processing to defend against adversarial attacks on hate speech detection models \cite{grondahl2018all,khieu2019tsar}. For instance, Moh et al. \cite{moh2020no} propose four different pre-processing defense techniques, namely {\em(i)} word segmentation no redo (WSNR), {\em(ii)} word balance no redo (WBNR), {\em(iii)} good grammar no redo, and {\em(iv)} vowel search no redo (VSNR). The defense techniques deal with white-space removal, typos, benign word insertion, and character boundary appending attacks, respectively. 

Table \ref{tab:soa_hate_speech_detection} summarizes some key papers on adversarial attacks and defense methods for hate speech detection.

%%%%%%%%%%%%%%%%%%%%%%%%%%%%%%%%%%%%%
 \begin{table}[]
\centering
% \resizebox{\textwidth}{!}{%
\scalebox{0.90}{
\setlength{\tabcolsep}{2.0pt}
\begin{tabular}{@{}lllll@{}}
\toprule
\multicolumn{1}{c}{\textbf{Ref.}} & \multicolumn{1}{c}{\textbf{Dataset}} & \multicolumn{1}{c}{\textbf{Method}} & \multicolumn{1}{c}{\textbf{Attack}} & \multicolumn{1}{c}{\textbf{Defense}} \\ \midrule

\cite{moh2020no}& \begin{tabular}[c]{@{}l@{}}W, T1, \\T2, \\T3 \cite{davidson2017automated} \end{tabular}&\begin{tabular}[c]{@{}l@{}} MLP, \\LA, \\ LSTM\end{tabular}& \begin{tabular}[c]{@{}l@{}} Pre-processing \\ attacks \end{tabular}& \begin{tabular}[c]{@{}l@{}}WSNR, \\WBNR, \\ GGNR, \\VSNR\end{tabular} \\ 
\cite{tran2020habertor}& \begin{tabular}[c]{@{}l@{}}Self \\collected\end{tabular} &BERT & 
\begin{tabular}[c]{@{}l@{}}Fast \\Gradient \\Method (FGM) \end{tabular}& \begin{tabular}[c]{@{}l@{}}Adversarial \\training \end{tabular}\\ 

\cite{grondahl2018all}&\begin{tabular}[c]{@{}l@{}}W, T1, \\T2, \\T3 \cite{davidson2017automated} \end{tabular} & \begin{tabular}[c]{@{}l@{}}LSTM, \\RNNs \end{tabular}&\begin{tabular}[c]{@{}l@{}}word \& \\word-boundary \\changes, \\appending \\unrelated \\words \end{tabular}& \begin{tabular}[c]{@{}l@{}}Transfer learning\end{tabular} \\ 
 \bottomrule
\end{tabular}%
}
\caption{Recent work of adversarial techniques for \textbf{hate speech detection}. MLP: Multilayer Perceptron, LA: Linear Regression, WSNR: word
segmentation no redo, WBNR: word balance no redo, GGNR: good grammar no redo, VSNR: vowel search no redo. Here T1, T2, T3 represent three different datasets collected from Twitter while W represents a dataset collected from Wikipedia. }
\label{tab:soa_hate_speech_detection}
\end{table}
%%%%%%%%%%%%%%%%%%%%%%%%%%%%%%%%%%%%%
\subsection{Misinformation}
\label{sec:misinformation}

%\st{Recent years showed real and large-scale adversarial attacks on social medial applications. A popular example is related to the announcement by Twitter in 2016 of blocking a large number of Bot/Troll accounts with adversarial attacks, largely to influence the US 2016 elections. 
%Machine-based text generation is not new. It has been used for several years in many applications such as customer services, marketing, etc. Those kinds of applications imply that users are aware that such text is created by machines, not humans. As such, misinformation in the scope of this paper, may not only arise from realizing that subject text is not accurate, but also where users are fooled to believe that subject text is created by humans and not machines. 
%Machine-based text generators described in the recent literature can be classified into different categories. In one category, we can look into those based on input text needed to the generator. For example, Generative Adversarial Network (GANs: \cite{goodfellow2014generative} can generate text from noise, OpenAI GPT models (e.g. \cite{radford2019language} can generate text given a simple context such as part of a sentence or a topic, models such as Grover \cite{zellers2019defending} can generate text based on causal language models, entities, etc.} 

Misinformation is perhaps the most innocent of the terms discussed here. Its misleading information is created or shared without the intent to manipulate people. An example would be sharing a rumor that a celebrity died, before finding out that it is false.

Disinformation, by contrast, refers to deliberate attempts to confuse or manipulate people with dishonest information.

%\fa{Misinformation refers to the information that is %false but not created to cause % harm to entity, person, organization, or society %\cite{ireton2018journalism}.
%This is illustrated by the definitions of these %notions by First %Draft,\footnote{\url{http://firstdraftnews.org/wp-co%ntent/uploads/2018/07/Types-of-Information-Disorder-%Venn-Diagram.png}} where %\emph{\textbf{misinformation}} is defined as %``\emph{unintentional mistakes such as inaccurate %photo captions, dates, statistics, translations, or %when satire is taken seriously}'' 
%}

To fight against such false or misleading information, several initiatives for manual fact-checking have been launched. Some notable fact-checking organizations include \emph{FactCheck.org},\footnote{\label{factcheck}\url{http://www.factcheck.org}}
\emph{Snopes},\footnote{\label{snopes}\url{http://www.snopes.com/fact-check/}}
\emph{PolitiFact},\footnote{\label{politifact}\url{http://www.politifact.com}}
and \emph{FullFact}.\footnote{\label{fullfact}\url{http://fullfact.org}} 
A large body of research focused on developing automatic systems for detecting the factuality of such information \cite{Li:2016:STD:2897350.2897352,Shu:2017:FND:3137597.3137600,Lazer1094,Vosoughi1146,vo2018rise}.

% In social networks, we can also analyze adversarial attacks from different perspectives.
Such detection systems are also vulnerable to adversarial attacks. \cite{zhou2019fake,guo2020future}.

The study by \cite{zhou2019fake} demonstrates three adversarial examples: Fact distortion, subject-object exchange and cause confounding. As a defense mechanism against adversarial attacks, the authors propose a crowdsourced knowledge graph to collect timely facts about news events. \cite{wu2019misinformation} describes adversarial attacks on social networks from two perspectives, attacks or manipulation of the network versus manipulation of the content. 

% FA: we cited this in the intro, it is not based on misinformation. \cite{szegedy2013intriguing} have shown that the trained deep learning models such as neural networks may fail to work against adversarial attacks. 

%\fa{please make sure the following papers are based on misinformation, for example, this one: \cite{akhtar2018threat}, on vision.}
%Those models can be very sensitive to changes and so small changes or perturbations to input vectors or features may cause such models to make incorrect predictions \cite{akhtar2018threat}. 

Several papers (e.g. \cite{guo2020future} called for the need to build adaptive ML models that are more immune and robust to variations and perturbations of text inputs or features.

Several defense approaches against adversarial attacks on NLP are proposed in the literature (e.g. adversarial training, \cite{miyato2016adversarial}, \cite{sato2018interpretable}, \cite{zhu2019freelb}, optimization-based methods: e.g. \cite{madry2017towards}, \cite{athalye2018obfuscated}, defense against neural fake news: e.g. \cite{zellers2019defending} and \cite{schuster2020limitations}, \cite{le2020malcom}, and word/sentence embedding-based defense: e.g. \cite{zhang2018word}, \cite{kula2019application}. \cite{qiu2019review} discussed three defense methods to adversarial attacks on misinformation: data modification, models modification and using auxiliary tools.

%%%%%%%%%%%%%%%%%%%%%%%%%%%%%%%%%%%%%
 \begin{table}[]
\centering
% \resizebox{\textwidth}{!}{%
\scalebox{0.78}{
\setlength{\tabcolsep}{2.0pt}
\begin{tabular}{@{}lllll@{}}
\toprule
\multicolumn{1}{c}{\textbf{Ref.}} & \multicolumn{1}{c}{\textbf{Dataset}} & \multicolumn{1}{c}{\textbf{Method}} & \multicolumn{1}{c}{\textbf{Attack}} & \multicolumn{1}{c}{\textbf{Defense}} \\ \midrule

\cite{miyato2016adversarial}&\begin{tabular}[c]{@{}l@{}}IMDB\\DBpedia\\Self \\collected\end{tabular} &LSTM &\begin{tabular}[c]{@{}l@{}} Random \\ perturbation\end{tabular} & \begin{tabular}[c]{@{}l@{}}Adversarial \\training \end{tabular}\\ 
\cite{kula2019application}& &\begin{tabular}[c]{@{}l@{}}RNN,\\ BERT \end{tabular}&\begin{tabular}[c]{@{}l@{}} Pre-processing \\ attacks\end{tabular} & \begin{tabular}[c]{@{}l@{}}Word/sentence \\embedding-based\\ defense\end{tabular} \\ 

\cite{le2020malcom}&\begin{tabular}[c]{@{}l@{}}GOSSIPCOP \cite{shu2018fakenewsnet},\\ PHEME \cite{kochkina2018all}\end{tabular} & RNNs& \begin{tabular}[c]{@{}l@{}}Comment generator \\in white box\\ setting\end{tabular}& \begin{tabular}[c]{@{}l@{}}Adversarial \\training\end{tabular} \\ 
%& & & & \\ 
%& & & & \\ 
%& & & & \\ 
 \bottomrule
\end{tabular}%
}
\caption{Recent work of adversarial techniques for \textbf{misinformation detection}.}
\label{tab:soa_mis_information_detection}
\end{table}
%%%%%%%%%%%%%%%%%%%%%%%%%%%%%%%%%%%%%
\subsection{Sentiment Analysis}
\label{sec:sentiment}

Sentiment analysis, which is also known as opinion mining, is another interesting application of NLP, and generally involves the use of modern technologies 
% such as ML and data analytics, along with NLP 
to analyze and extract quantitative results.
% from multimedia content. 
Generally, the results of sentiment analysis are presented in the form of \textit{positive}, \textit{negative}, and \textit{neutral} sentiments. 
% FA: let us not mix with emotion, However, other categories and labels, such as \textit{angry}, \textit{happy}, and \textit{sad} etc or \textit{interested} and \textit{not interested}, depending on the application could also be used. 
In the literature, both textual and visual contents have been analyzed to extract opinions about an entity \cite{hassan2019sentiment,hassan2020visual,alfarrarjeh2017geo}. However, text is more exploited due to a diversified set of applications and the level and freedom of expressing sentiments %and emotions in text \cite{birjali2021comprehensive}. 
% FA: let's not talk about applications as do not mention applications in other sections 
%Some of the key applications of textual sentiment analysis include brand monitoring and reputation management \cite{benedetto2016big}, customer support and feedback \cite{shakhovska2020sentiment}, market research \cite{nguyen2015deep}, natural disaster analysis \cite{beigi2016overview}, and teacher evaluation \cite{rajput2016lexicon} etc. 

The recent increase in the popularity of social media outlets, such as Twitter and Facebook, has further increased the importance, opportunities, and challenges associated with sentiment analysis \cite{ahmed2015sentiment}. For instance, sentiment analysis tools allow businesses to monitor and analyze the popularity and users' feedback on them and their competitors' products/services by processing text reviews shared by the users in online social networks. The objective quantitative results obtained through sentiment analysis are then utilized in making critical business decisions. 

Similar to other NLP applications, due to the importance of results obtained from sentiment analysis in the decision-making process, sentiment analysis algorithms are also subject to several adversarial attacks. In sentiment analysis, an attacker can launch adversarial attacks by adding small perturbations to text to generate different perceptions than the actual opinions. For instance, in \cite{de2021adversarial}, the vulnerabilities of a lexical natural language sentiment analysis algorithm are identified and analyzed under various types of adversarial attacks. Based on the experimental results, the authors conclude that the classifier's results could be significantly affected by exploiting the identified vulnerabilities. Alzntot et al. \cite{alzantot2018generating} on the other hand utilize a black-box population-based optimization algorithm for the generation of semantically and syntactically similar adversarial examples to launch attacks on sentiment analysis models. In \cite{hofer2021adversarial}, three different types of character-level adversarial attacks are launched against a BERT-based sentiment analysis framework. The strategies used for generating the adversarial examples include mimicking human behavior and using leetspeak, misspellings, or misplaced commas. These strategies allow maximizing misclassification rates of sentiment analysis classifier with minimal changes.

The literature also reports some interesting solutions to guard against adversarial attacks and extract correct perceptions on content shared in social networks. For instance, in \cite{karimi2021adversarial}, an adversarial training-based solution has been proposed for aspect-based sentiment analysis. To this aim, two different BERT models, pre-trained on general-purpose and domain-specific data, are fine-tuned in a novel framework-based adversarial training. Wang et al. \cite{wang2021textfirewall} on the other hand propose an adversarial defense tool namely \textit{TextFirewall} for sentiment analysis algorithms. \textit{TextFirewall} mainly relies on the inconsistency between the sentiment analysis model's prediction and the impact value, which is calculated by quantifying the positive and negative impact of a word on the sentiment polarity. Hosseini et al. \cite{hosseini2017deceiving} used spell-checking for limiting adversarial modifications at the character level. This scheme is also used in \cite{de2021adversarial} to guard a sentiment analysis classifier against insertion and word substitution attacks. Du et al. \cite{du2020adversarial}, on the other hand, employ network distillation technique along with adversarial training for robust sentiment analysis.

Table \ref{tab:soa_sentiment_analysis} summarizes some key findings on adversarial attacks and defense methods for sentiment analysis.

%%%%%%%%%%%%%%%%%%%%%%%%%%%%%%%%%%%%%
 \begin{table}[]
\centering
% \resizebox{\textwidth}{!}{%
\scalebox{0.78}{
\setlength{\tabcolsep}{2.0pt}
\begin{tabular}{@{}lllll@{}}
\toprule
\multicolumn{1}{c}{\textbf{Ref.}} & \multicolumn{1}{c}{\textbf{Dataset}} & \multicolumn{1}{c}{\textbf{Method}} & \multicolumn{1}{c}{\textbf{Attack}} & \multicolumn{1}{c}{\textbf{Defense}} \\ \midrule

\cite{wang2021textfirewall}&\begin{tabular}[c]{@{}l@{}}IMDB \cite{maas2011learning}, \\Yelp \cite{zhang2015character} \end{tabular}& DNNs& \begin{tabular}[c]{@{}l@{}}Deepwordbug \cite{gao2018black}, \\GA \cite{alzantot2018generating}, \\PWWS \cite{ren2019generating}\end{tabular}& \begin{tabular}[c]{@{}l@{}}Considers inconsistency \\between the model’s \\output and the
\\impact value.\end{tabular} \\ 
\cite{alzantot2018generating}&IMDB &DNNs &\begin{tabular}[c]{@{}l@{}}black-box \\population-based\\ optimization \end{tabular}& Adversarial Training \\ 

\cite{de2021adversarial}&IMDB &\begin{tabular}[c]{@{}l@{}}lexical \\natural language \\classifier \end{tabular}& \begin{tabular}[c]{@{}l@{}}Characters insertion \\Word substitution\end{tabular}& N/A \\ 
 \bottomrule
\end{tabular}%
}
\caption{Recent work of adversarial techniques for \textbf{sentiment analysis}.}
\label{tab:soa_sentiment_analysis}
\end{table}
%%%%%%%%%%%%%%%%%%%%%%%%%%%%%%%%%%%%%

\section{Research Challenges, and Lesson Learned} % and Future Research Directions}
\label{sec:challenges}

%\subsection{Challenges and Open Issues}
% and Future Research Trends
\subsection{Research Challenges} 
% \par After reviewing the main contributions achieved in the development of the understanding of adversarial attacks and defense, it is time to highlight some existing challenges that need to be addressed and alleviated for the advancement of this research area. In essence, the majority of such challenges originate from what distinguishes text from image data, since the majority of adversarial security work has been conducted so far from a computer vision/image processing perceptive, and less work considers text. Therefore, we revise those challenges keeping in mind the unique limitations one faces when dealing with text. 
% Here is a detailed account of outstanding 
In this section, we discuss current challenges and limitations in the area of adversarial attack generation and defense in text-based social media applications. 

\begin{itemize}[leftmargin=*]

\item \textbf{Discrete data perturbation}: This is a fundamental challenge in generating adversarial examples in a text as compared to images. 
% the early usage of ML models in computer vision where the input is continuous. 
It is widely believed that computer vision and image processing approaches to adversarial example generation do not directly apply to discrete inputs. In essence, such a direct application will result in exposing the attack as characters and words are not salient anymore. This is the case as perturbations have to be discrete. Moreover, text perturbations are mainly replacement operations. This poses an important research question on \textit{how to characterize good replacements?}. Hence, an active area of research considers identifying and quantifying new aspects of text similarity.
Similar to attack methods, continuous input defense techniques cannot be easily applied. As an example, the GAN approach is based on adding artificial noise to the inputs. Thus, it is not applicable in an NLP context. Accordingly, an interesting research area is how to enable the off-the-shelf adversarial generation and defense methods of continuous variables to be used in a textual context. 
 
\item \textbf{Discrete data- perceivability}: While small arbitrary image perturbations are typically imperceptible to the human eye, humans can easily detect text modifications. This poses certain limitations on what makes a text modification imperceptible to human understanding. Recent research highlights grammatical, semantic, and context correctness \cite{pruthi2019combating} for viable adversarial text replacements. Still, imposing such strict constraints is known to restrict the space of possible replacements, thereby degrading the strength of the generative adversarial examples. Given this trade-off, there is an immense need for more research on characterizing what defines an imperceptible, yet efficient adversarial example.
 
%\textcolor{blue}{ FA: The following point seems very general. I would skip this point.} 
\item \textbf{Human-factor dependency:} As seen in the papers surveyed in this section, there is almost always a human effort in the design and optimization of adversarial attack and defense methodologies. This is especially the case for measuring the imperceptibility of adversarial replacements from a linguistic point of view \cite{ribeiro2018semantically}. Other human factor example aspects include concerns with setting the parameters of the attack model and optimizing such parameters. Therefore, it seems interesting to invest more research on how to automate these processes in an attempt towards eliminating or, at least, regulating the extent to which the human factor is necessary for this area \cite{liang2017deep}. In fact, this is a general research challenge that concerns ML in general, including the context of this section. 
 
\item \textbf{Transferability- of attack and defense}: Similar to the case in computer vision, adversarial examples in an NLP context are known to possess different aspects and extensions of transferability. Typically, they transfer across different models as well as training and testing datasets. Furthermore, recent research envisions focused on developing attacks that transfer even across different ML tasks, as well \cite{demontis2019adversarial}. The transferability character exempts an attacker from the need to use the same model, architecture, data, or even a task of the attacked ML model \cite{wallace2019universal}. This resembles a growing challenge against efficient defense techniques \cite{le2020malcom}. As intuitively expected, black-box and untargeted adversarial attacks have higher degrees of transferability as compared to white-box and/or targeted attacks. Therefore, there will be an increased demand for developing advanced defense techniques that work with this challenge. It is noted that transferability can be treated at the character, word, and sentence levels. 

 \item \textbf{Universality- of attack and defense:} This property refers to the extent to which an adversary is independent of the input, either status of nature. Thus, in an NLP context, the universality of an adversary can be seen in how it can attack irrespective of the input. For example, an NLP attack can be independent of the language used. Another aspect is defense universality; the robustness against character-, word-, and sentence-level attacks at the same time. The incorporation of this property in adversarial attack and defense is an interesting area of research. Recent research signifies that the success of a defense strategy against a specific adversarial attack may not necessarily mean the system is robust against other adversaries. Therefore, developing universal defense mechanisms is of growing importance. This is especially the case with the state-of-the-art deep learning models that can learn the task irrespective of inputs or architecture. Besides, it is also worth investigating the relationship between efficient defenses against different attacks, and then the attacker-attacker relationship holds between their respective defenders.
 
 \item \textbf{Lack of appropriate defense:} 
% The research body in the adversarial security literature necessities the need for efficient defense techniques. Besides, such techniques should define the model for a variety of possible unknown adversarial attacks. 
 An area worth investigating is whether defense methods have the transferability property, and if so, in what sense? and, more importantly, how to enhance and boost this transferability to have a generic defense approach? 
\par Improved detection of grammatical errors is worth a further investigation \cite{wallace2019universal,jin2020bert}. This will have the added benefit of decreasing the chances of adversarial attacks. 
\par The development of more robust NLP architectures: Recent research works hint at a promising research outlook, which is on how to design inherently robust/defensive ML architectures that can suppress the impact of any adversarial attack \cite{belinkov2017synthetic}.

\item \textbf{Embedding space size-perturbation selection trade-offs}: Several works on characterizing a suitable embedding space have been proposed. Besides, there is a variety of methods to search for the best replacement amongst the elements in the embedding space. Still, there is a need for more work on how to define the space, and how to efficiently search between its elements approaching the best replacement. This requires also balancing the trade-off between the embedding space size and the search computational burden. Moreover, there is a need for novel ideas on how to regularize and guide the search process within the embedding space. A few works have considered the question of identifying critical characters and words that are best to change amongst all others. However, there is still an evident need for more research on this idea. In this context, one may think of developing a unified framework for answering a more common question of what, where, when, and how to change, i.e., is it better to change a character, as (sub) word or a (sub) sentence? This opens up the horizon for developing new optimization settings that may prove to answer such questions in a systematic or tractable way.

%\textcolor{blue}{FA: I am not able to imagine what type of dataset we specifically need. For NLP and social media applications, there are plenty of datasets, and in my understanding, all datasets can be used for adversarial study..} \textcolor{red}{I think they mean a collection of adversarial example}
 \item \textbf{Lack of standards in benchmark datasets with proper collection of adversarial examples or instances :} Given the fact that there is a variety of datasets used in training, testing, and performance evaluation, there is a need for standardizing benchmark datasets. Although several large-scale datasets have been collected for common NLP tasks in recent years, there remains a need for new datasets for more challenging social networks NLP applications. It is noted that this is a general concern in text and NLP adversarial attack research areas.

\item \textbf{Lack of standards quality metrics:}
With the diversity of attack and defense methods, there is a corresponding diversity of employed attack and defense quality metrics. A classical attack quality metric is the accuracy of the targeted ML model. Furthermore, in an NLP context, the semantic similarity between the original input and its adversarial replacement has also been widely used as a quality metric. Along this line, researchers have used Blue, Self-Blue, EmbSim, Euclidean distance, cosine, and semantic similarity to evaluate their attack methods or the quality of a generated text. On the other hand, human evaluation seems to be an unavoidable measure of imperceptibility \cite{ribeiro2018semantically}. To this end, there is an immense need for research on developing descriptive, fair, and standardized performance evaluation metrics for the quality of the attack and the defense. Moreover, other quality metrics evaluating other aspects of the attack, such as transferability, universality, and imperceptibility need also to be developed and evaluated.
From an adversarial replacement generation perspective, there is a gowning need for standardizing measures and rules for semantic similarity. This is of great importance for ensuring the imperceptibility of attacks. Moreover, there is a need for developing and standardizing measures for the ML model's stability against adversarial attacks. Along the line of quality metrics, \cite{michel2019evaluation} recently showed 
%necessitate 
the added benefit of incorporating the attack's meaning preservation capability into quality metrics for attack performance evaluation. Also, a core question is on defining the distance between the original text and its adversarial modification, i.e., on \textit{how to measure a text change?}

 \item \textbf{Computational complexity concerns:} Either on the attack or defense sides, there is an inherent trade-off between the performance and computational complexity, just like any other application area. It can be seen that settings with more access to system parameters are better in performance but at the cost of more computational complexity. With The increase of the attack and defense degrees of complexity, there should be an accompanying research effort on how to implement such techniques at affordable computational complexity levels, execution time, and memory requirements. Similar to several aforementioned research challenges, computational complexity is a general concern in ML in general, including the context of this section.
 
 \item \textbf{Balance between protection against attacks, censorship, and freedom of speech:} Social media outlets and feeds from these outlets can be attacked with groups with radical views to promote their views. Thus, the social network applications are more susceptible to attacks and mitigation techniques are hard to integrate with social platforms as that might be seen as an attack on ``freedom of speech.'' It is important to maintain the balance between protection against attacks, censorship, and freedom of speech.

% Graph Neural Networks: is a new item added by Mahmoud Nazzal. Please read and correct.
% I have updated Fig. 6 accordingly. 
\item \textbf{Graph Neural Networks:} 
Graphs representation has been receiving a lot of attention as a key framework for representing entities with relations. In the context of a social media network, users, user information, and relationships between users can be represented by graph nodes, node attributes, and graph edges, respectively. Besides, graph-level operations represent operations on the level of a social media network. As a typical example, predicting the link between two graph nodes corresponds to predicting a friendship or following relationship between two members on a social media platform \cite{gong2014joint}. As another example, estimating whether a node belongs to a subgroup corresponds to estimating whether a user belongs to a social media group. 

The primary added benefit of a graph representation is rendering the inherent links between nodes. Classical methods such as random walk can preserve these links. Nevertheless, the success of deep neural networks in various fields has led to the development of graph neural networks (GNNS), a special type of neural network well-suited for embracing node links in their representation across the network layers. GNNs have been receiving increasing attention in recent years in the representation of data with relationships. A key usage of GNNs is in the representation of social media networks. This elegant representation has tweaked downstream tasks such as node and graph classification and link prediction. 

Despite their success in modeling graph data, GNNs have been shown vulnerable to adversarial attacks \cite{zhang2020gnnguard}. In this context, imperceptible perturbations applied at the node, edge, graph, and attribute level are shown to trick the GCN. Therefore, reaping the promising potential of GNNs in representing social media networks requires improving their robustness to such attacks and preserving the privacy of the data they represent. Here is a summary of key research challenges facing the security and privacy of GNNs in the context of social media platforms.

\begin{itemize}[leftmargin=1em]
 \item \textbf{Addressing the dynamic nature of social media network graphs:} The majority of existing research considers static graphs with node attributes \cite{jin2021adversarial}. Still, social media networks are continuously dynamic \cite{kazemi2020representation}. For example, a person may make new friendships or terminate existing ones with certain persons. Also, a new indecent may happen and attract the interest of people all of a sudden, such as the so-called \textit{trending} events. Thus, this dynamic nature needs to be modeled well in GCNs and taken care of in the underlying design of adversarial attack and defense techniques. 
 \item \textbf{Imperceptibility-definition and quantification:} There is a need for defining to which extent a given perturbation is imperceptible. This requires developing novel imperceptibility metrics.
 \item \textbf{Identifying what makes a good perturbation:} An interesting research direction is to study the commonality between good attacks attained by the existing techniques. This can hint at how to systematically design good attacks. This will also be reflected on how to exploit this knowledge in designing efficient defense measures.
 \item \textbf{Extension of knowledge from other domains:} A possible research direction is to consider extending the existing techniques for attack and defense in text and NLP domains to the graph domain. 
 \item \textbf{Scalability of attack and defense methods:} there is a need for developing effective attack and defense methods at a small scale. Then, they can be easily scaled up to the level of the entire network.
 \item \textbf{Mitigating privacy leakage in GCNs:} Social media networks typically contain private data that should be kept away from the reach of the public, as well as publically published data; that the persons and platforms share publically at no harm. To this end, privacy leakage happens when public information can be exploited to infer private data. Examples along this line include revealing private links between persons based on public information \cite{korolova2008link}, and even re-identifying anonymized persons in a social media network based on public information \cite{narayanan2009anonymizing}. In fact, a series of recent works accuse GCNs to leak private data, especially in the context of social media \cite{duddu2020quantifying, he2021node}. The main reason for this leakage is the correlation between data of different users in a social GNN \cite{kifer2011no}. Thus, an important future research venue is to mitigate this leakage. Examples along this line include leveraging differential privacy and developing new GCN architectures to minimize this leakage. 
 
\end{itemize}
\end{itemize}

\begin{figure*}[t]
\centering
% \resizebox{0.98\columnwidth}{!}{
\includegraphics[width=0.99\textwidth]{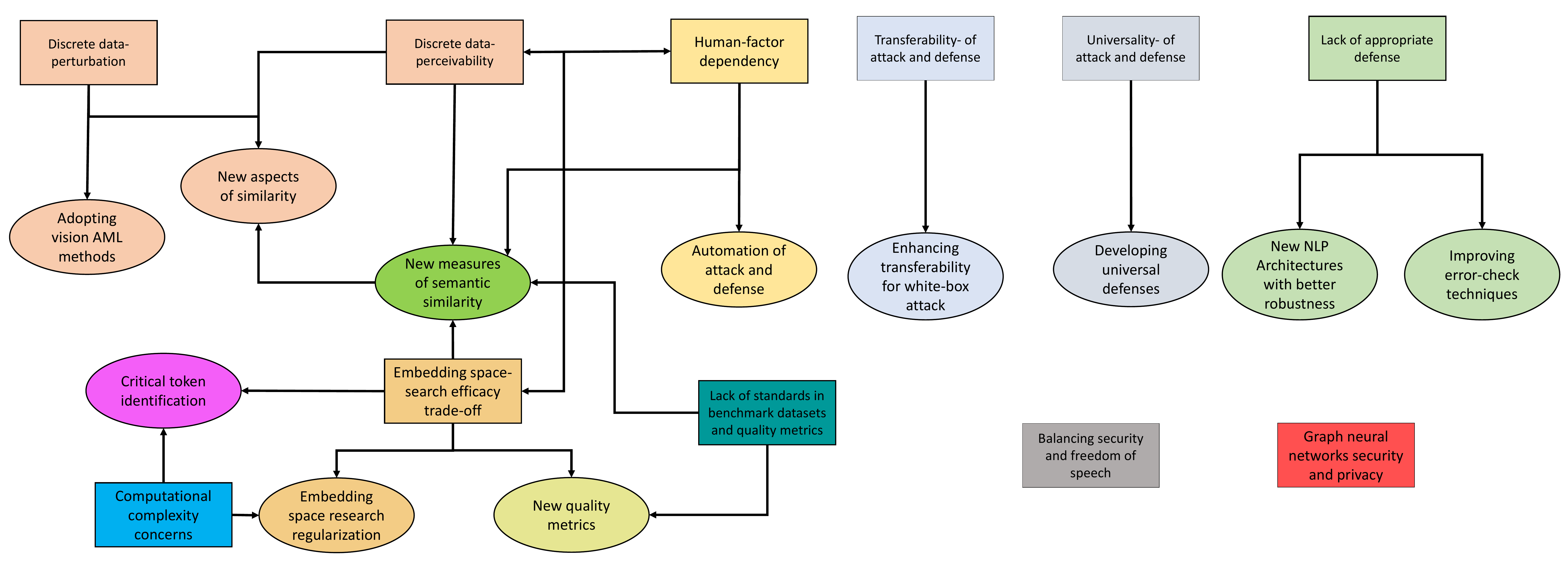}
% }
\caption{A summary of the main outstanding challenges in text adversarial attacks and defense research, along with relevant research trends and recommendations. Challenges are represented with rectangles, whereas research trends and recommendations are in oval shapes.}
\label{Challenges_and_trends}
\end{figure*}

%\textcolor{blue}{FA: i do not feel the need a figure here. Moreover, it is hard to follow.} \textcolor{red}{i agree}
Fig. \ref{Challenges_and_trends} summarizes the outstanding challenges in the areas of attack and defense, along with the underlying research trends; concurrent and future ones. 

\subsection{Discussion and Lessons Learned} 
\label{ssec:discussion}
%Online social networks data/text is different from the conventional sources of text, such as newspapers, magazine articles, books, surveys, and other types of question answering materials. Text shared in these networks is generally non-structured and can be written by different people including experts, non-experts, and non-professional writers in different languages and styles. Due to these characteristics processing social media content is more challenging. 

\begin{itemize}[leftmargin=*]
 \item Adversarial ML for social media NLP applications is a very important and growing research area as the stake is very high. For example, it allows influencing public opinions by hackers/attackers are typically from state-sponsor agencies.
 
 % Mahmoud: The last statement is not clear
 
 \item Textual content in social media is very noisy, 
 % Online social networks data/text, which is unstructured and written in a different dialect, languages, styles, and native language text are written in ASCII/English characters.
 %by different people including non-professional writers in different languages and styles. 
 Due to such diverse characteristics, NLP in general and adversarial NLP, in particular, is more challenging. Developing defense techniques is even harder. 
 \item Online social media applications are more prone to adversarial attacks than conventional sources of text.
 \item Social media applications are more susceptible to attacks and mitigation techniques are hard to integrate with social platforms as that might be seen as an attack on ``freedom of speech.'' So there is a delicate balance between protection against attacks, censorship, and freedom of speech. 
 \item The widespread use of social media attracts the attackers to launch different types of adversarial attacks on these networks to fulfill their objectives. 
 \item Social media outlets are distributed in nature, therefore, distributed attacks are easier to manifest in social media applications. 
 \item A successful attack against the most vulnerable social media outlets might be enough to have unintended consequences (e.g., promote extreme views, etc.).
 \item The unstructured nature of text shared in social networks allows attackers to launch different types of adversarial attacks on the applications. 
 \item The literature indicates that several interesting NLP applications of social networks are subject to adversarial attacks. Some of the notable applications include rumors, satires, parodies, clickbait, spam, hate speech, and misinformation detection. 
 \item Recent literature necessities the need for developing efficient adversarial defense techniques.
 \item There is a need for reducing human dependency both in adversarial attack and defense
 \item The existing literature on adversarial NLP in general and in social media applications in particular lacks in focused benchmark datasets and quality metrics.
 \item Graph neural networks form a promising framework with a great potential for improving the representation of social media information. however, this requires improving their security against adversarial attacks and mitigating their privacy leakage. 
\end{itemize}
\section{Conclusions}
\label{conclusion}
We surveyed the state-of-the-art in adversarial attack and defense on \textit{five} major text-based social media applications, which include rumors, satires, clickbait \& spam, hate speech, misinformation detection, and sentiment. These applications are susceptible to adversarial attacks. In this paper, we first provide a general overview of adversarial attack and defense techniques applicable to both text and image applications. As text is a primary communication means in social media platforms, we then review state-of-the-art techniques for attack and defense techniques specifically applicable for NLP. After that, we discuss the attack and defense studies in the aforementioned five major social media applications. Finally, we highlight current research challenges in the context of security of ML for social media applications and state key lessons.

% The spread of misinformation through social media in recent years is overwhelming. Our goal in this paper is to help researchers learn about some of the challenges, open issues, research trends, and challenges in this area. 

% We evaluated in this paper different aspects of misinformation especially in (OSNs) such as rumors, clickbaits, spams, etc. The types and volume of misinformation will continue to expand in OSNs due to the nature of such networks. OSNs are open, the process of creating users and/or content is largely not vetted or checked. 
% Another major problem is related to the nature of misinformation, especially when it crosses subjective areas such as politics, religions, etc. 

% We differentiated in this paper, between genuine users in OSNs spreading misinformation, knowingly or unknowingly, versus adversarial attackers creating bots and trolls in OSNs and masquerading those accounts as genuine users. How can we judge whether a text generated in OSNs is adversarial or not? How can we distinguish a text created by a human, versus a text created by a bot or troll? 

\section*{Acknowledgment}
%This publication was made possible by NPRP grant \# [13S-0206-200273] from the Qatar National Research Fund (a member of Qatar Foundation). The statements made herein are solely the responsibility of the authors.
The authors extend their appreciation to the Deputyship
for Research \& Innovation, Ministry of Education in Saudi
Arabia for funding this research work through the project
number 1120.
\bibliography{elsarticle-template}

\end{document}